\documentclass[sigconf, nonacm]{acmart}

\usepackage{amsmath,amsfonts}
\usepackage{pifont}
\usepackage{mathtools}
\usepackage{hhline}
\usepackage[linesnumbered,ruled,vlined]{algorithm2e}
\usepackage{empheq}
\usepackage{caption}
\usepackage{wrapfig}
\usepackage{cases} 
\usepackage{float}
\usepackage[misc]{ifsym}
\usepackage{bbding}
\usepackage{fontawesome}
\usepackage{microtype}
\usepackage{subcaption}
\usepackage{booktabs}
\usepackage{multirow}
\usepackage[switch]{lineno}
\usepackage{textcomp}
\usepackage{multicol}
\PassOptionsToPackage{dvipsnames}{xcolor}
\usepackage{pgfplots,tikz}

\pgfplotsset{compat=1.16}
\usetikzlibrary{shapes,arrows,positioning,fit,
backgrounds,calc,patterns,hobby,matrix}
\def\BibTeX{{\rm B\kern-.05em{\sc i\kern-.025em b}\kern-.08em
    T\kern-.1667em\lower.7ex\hbox{E}\kern-.125emX}}
\usepackage{array}
\usepackage{etoolbox}
\usepackage{hyperref}
\usepackage{url}

\usepackage[normalem]{ulem}
\newcommand{\cmark}{\ding{51}}%
\newcommand{\xmark}{\ding{55}}%
\newcolumntype{P}[1]{>{\centering\arraybackslash}p{#1}}
\SetKwInput{KwInput}{Input}                
\SetKwInput{KwOutput}{Output}              
\SetKw{Break}{break}

\newcommand\vldbdoi{10.14778/3494124.3494142}

\newcommand\vldbavailabilityurl{https://github.com/d-gcc/CAE-Ensemble}
\newcommand\vldbpagestyle{empty}

\begin{document}

\title{
Unsupervised Time Series Outlier Detection with Diversity-Driven 
Convolutional Ensembles---Extended Version} 

\author{David Campos$^{1*}$, Tung Kieu$^{1*}$, Chenjuan Guo$^{1+}$, Feiteng Huang$^2$, Kai Zheng$^3$, Bin Yang$^1$, and Christian S. Jensen$^1$}

\affiliation{%
  \institution{$^1$Aalborg University, Denmark $^2$Huawei Cloud Database Innovation Lab, China\\ $^3$University of Electronic Science and Technology of China, China}
  \country{$^1$\{dgcc, tungkvt, cguo, byang, csj\}@cs.aau.dk, $^2$huangfeiteng@huawei.com, $^3$zhengkai@uestc.edu.cn} \\
}







\begin{abstract}
With the sweeping digitalization of societal, medical, industrial, and scientific processes, sensing technologies are being deployed that produce increasing volumes of time series data, thus fueling a plethora of new or improved applications. In this setting, outlier detection is frequently important, and while solutions based on neural networks exist, they leave room for improvement in terms of both accuracy and efficiency.
With the objective of achieving such improvements, we propose a diversity-driven, convolutional ensemble. To improve accuracy, the ensemble employs multiple basic outlier detection models built on convolutional sequence-to-sequence autoencoders that can capture temporal dependencies in time series. Further, a novel diversity-driven training method maintains diversity among the basic models, with the aim of improving the ensemble’s accuracy. To improve efficiency, the approach enables a high degree of parallelism during training. In addition, it is able to transfer some model parameters from one basic model to another, which reduces training time. 
We report on extensive experiments using real-world multivariate time series that offer insight into the design choices underlying the new approach and offer evidence that it is capable of improved accuracy and efficiency. 

This is an extended version of ``Unsupervised Time Series Outlier Detection with Diversity-Driven
Convolutional Ensembles''~\cite{pvldb22}, to appear in PVLDB 2022.

\end{abstract}
\renewcommand*{\authors}{David Campos, Tung Kieu, Chenjuan Guo, Feiteng Huang, Kai Zheng, Bin Yang, and Christian S. Jensen}
\maketitle

\pagestyle{\vldbpagestyle}
\begingroup
 \renewcommand\thefootnote{}\footnote{\noindent
 $*$: Equal contributions. +: Corresponding author. \\ 
 }
\addtocounter{footnote}{-1}\endgroup





\section{Introduction} 
\label{introduction}


As part of the continued digitization, processes are increasingly being instrumented with sensors, which offer  
monitoring capabilities with the objective of supporting a variety of applications~\cite{DBLP:conf/icde/LiuJYZ18,Hundman18,DBLP:conf/ijcai/YangGHT021,DBLP:conf/icde/Pedersen0J20,DBLP:conf/icde/Yang020}. 
%
The monitoring of processes results in time series data~\cite{razvanicde2021,DBLP:journals/vldb/HuYGJ18}. 
%
In many cases, it is important to be able to use this data to identify outliers or anomalies~\cite{Aggarwal13}. Outliers are relatively uncommon observations that differ 
from the remaining observations in a series. The accurate detection of outliers is critical in applications such as health, web,  manufacturing, and transportation~\cite{DBLP:journals/vldb/PedersenYJ20,DBLP:journals/vldb/GuoYHJC20,DBLP:journals/pvldb/PedersenYJ202,DBLP:conf/waim/YuanSWYZY10}. 

We consider \emph{unsupervised} outlier detection, meaning that we do not rely on outlier labeled data for training, which offers two benefits. First, labeling by human experts is time consuming and labor intensive; thus, time series often come without outlier labels. Second, the unsupervised setting eliminates the reliance on manual labeling, where some outliers may go unnoticed, thus potentially enabling the detection of unanticipated types of outliers.

Unsupervised outlier detection has traditionally been supported by linear methods~\cite{Gupta14} that require human experts to set model parameters for the specific scenarios. The recent use of deep learning techniques enables substantial detection accuracy improvements over the linear models because they are able to learn non-linear features from the data, such as complex temporal dependencies, without requiring explicit supervision by human experts. 

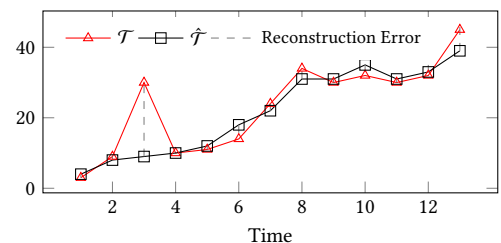
\begin{figure}[ht!]
\small
  \centering
  \begin{tikzpicture}
    \begin{axis}[
        xlabel=Time,
        width=0.9*\linewidth,
        height=0.55*\axisdefaultheight,
        xtick={2,4,6,8,10,12},
        legend pos= north west, legend columns=-1,
        legend style={draw=none}]
        \addplot[red, mark=triangle] table[x=t, y=b] {figures/data/series.txt};
        \addlegendentry{{\footnotesize $\mathcal{T}$}}
        \addplot[black, mark=square] table[x=t, y=a] {figures/data/series.txt};
        \addlegendentry{{\footnotesize ${\hat{\mathcal{T}}}$}}
        \addplot[gray, dashed, quiver={v=\thisrow{c}}] table[x=t, y=a] {figures/data/series.txt};
        \addlegendentry{{\footnotesize Reconstruction Error}}
        \end{axis}
    \end{tikzpicture}
  \caption{{\texttt{AE}-based unsupervised outlier detection.
  }} 
  \label{fig:reconstruction}
\end{figure}

Autoencoders (\texttt{AE})~\cite{Hawkins02,DBLP:conf/cikm/Kieu0GJ18} represent a popular and powerful deep learning based approach to unsupervised outlier detection. An \texttt{AE} consists of an \emph{encoding} phase that compresses an original time series $\mathcal{T}$ into a compact representation and a subsequent \emph{decoding} phase that reconstructs an output time series $\hat{\mathcal{T}}$ from the compact representation. 
The compact representation is only able to capture representative patterns that reflect general patterns in the original time series. 
As a result, the reconstructed time series $\hat{\mathcal{T}}$ is unlikely to capture outliers that by virtue of being outliers are not representative. Thus, if the difference between observations in $\mathcal{T}$ and in $\hat{\mathcal{T}}$, called the reconstruction error, exceeds a threshold $\epsilon$, this suggests an outlier. 
%
For example, this occurs at 
time 3 in
Figure~\ref{fig:reconstruction}.   

An autoencoder ensemble (\texttt{AE-Ensemble}) employs multiple \texttt{AE}s to avoid a single \texttt{AE} being overfit to original data~\cite{Chen17, Okun11}, which often further improves accuracy. 
In particular, \texttt{RAE-Ensemble} achieve high accuracy using an ensemble of recurrent neural network based autoencoders (\texttt{RAE}s)~\cite{Kieu19}, where a \texttt{RAE} is able to capture temporal dependencies among the observations in time series~\cite{Malhotra16,Kieu18},
e.g., the increasing trend shown in Figure~\ref{fig:reconstruction}. 
This is a highly desirable property compared to classic \texttt{AEs} that use, e.g., feedforward neural networks, which fail to capture temporal dependencies. 



However, \texttt{RAE-Ensemble} may still be improved in terms of accuracy and efficiency. 
First, while an ensemble often benefits from diverse basic models~\cite{Aggarwal17}, 
the basic models in \texttt{RAE-Ensemble} may not be sufficiently diverse. 
Specifically, \texttt{RAE-Ensemble} generates basic models with randomly different network structures by randomly adding skip connections. However, the resulting basic models may not necessarily return diverse results, 
offering room for higher accuracy. 
%
Second, the basic models in \texttt{RAE-Ensemble} are based on Recurrent  Neural  Networks (\texttt{RNN}s), which are \emph{inefficient}, 
as an \texttt{RNN} must process the time series data sequentially, as opposed to in parallel, due to its recurrence nature.


We propose \texttt{CAE-Ensemble}, a novel autoencoder ensemble approach that uses convolutional neural networks and aims to improve accuracy and efficiency by overcoming the limitations of \texttt{RAE-Ensemble}. 
First, we introduce a diversity metric that makes possible to assess the suitability of a newly generated basic model given the current state of the ensemble. This ensures the diversity among the basic models. 
%
Second, we replace the \texttt{RNN} framework of \texttt{RAE-Ensemble} with Convolutional Neural Networks (\texttt{CNN})~\cite{LeCun98,Gehring17}, in order to capture temporal dependencies, with the benefit of achieving a high degree of training parallelism, thus improving efficiency.
%
Third, to further reduce the training time, \texttt{CAE-Ensemble} integrates a transfer learning method based on so-called born-again networks~\cite{Furlanello18} that share a portion of the trained parameters from one basic model with other basic models. 
Fourth, we propose a fully unsupervised strategy for selecting hyperparameters without relying on ground truth outlier labels. 
%
%

To summarize, we make the following contributions. 
First, we propose a convolutional sequence-to-sequence autoencoder \texttt{CAE} for outlier detection that is capable to capture temporal dependencies with a high degree of training parallelism. Second, to improve the accuracy and efficiency, we propose a diversity-driven ensemble using \texttt{CAE}s as its basic models, along with a parameter transfer based training strategy. Third, we propose a fully unsupervised strategy to select hyper-parameters, and finally, we report insights from extensive experiments with real-world time series data sets to justify design choices in our proposal.


The paper is organized as follows. 
Section~\ref{definitions} 
formalizes the problem. 
Section~\ref{framework} elaborates 
\texttt{CAE-Ensemble}.    
Section~\ref{experiments} reports on the experiments. 
Section~\ref{relatedwork} reviews related work. Section~\ref{conclusion} concludes. 

\section{Preliminaries} \label{definitions}

\textbf{Time Series Outlier Detection. }
A time series $\mathcal{T} = \langle \mathbf{s}_{1}, \mathbf{s}_{2}, \dots, \mathbf{s}_{C} \rangle$ is a sequence of $C$ observations, where each 
observation $\mathbf{s}_{t} \in \mathbb{R}^D$ is a $D$-dimensional vector. If $D=1$, $\mathcal{T}$ is \textit{univariate}, and if $D>1$, $\mathcal{T}$ is \textit{multivariate}, or \textit{multidimensional}. 
%
Given a time series $\mathcal{T} = \langle \mathbf{s}_{1}, \mathbf{s}_{2}, \dots, \mathbf{s}_{C} \rangle$, the outlier score $\mathcal{OS}(\mathbf{s}_{t})$ for observation $\mathbf{s}_{t}$ is a value such that the higher $\mathcal{OS}(\mathbf{s}_{t})$ is, the more likely it is that $\mathbf{s}_{t}$ is an outlier.
Given an outlier score threshold $\epsilon$, outliers are the observations in $\mathcal{T}$ whose outlier score exceeds $\epsilon$. 
Usually, threshold $\epsilon$ is defined according to domain knowledge. 

In the experiments, to evaluate accuracy in a fair and meaningful manner, we use different metrics, including metrics that (1) require such specific thresholds and (2) do not rely on thresholds, e.g., when domain knowledge on selecting such thresholds is unavailable.  




\noindent
\textbf{Autoencoders. }
A classic autoencoder \texttt{AE} consists of an encoder and a decoder, where each is a feed-forward neural network. 
The encoder transforms a $D$-dimensional input $\mathbf{x}$ to an intermediate and compact vector $\mathbf{h} \in {R}^{M}$, where $M < D$. Then, the decoder transforms the intermediate vector $\mathbf{h}$ 
to an output vector $\hat{\mathbf{x}} \in {R}^{D}$ that approximates the input vector. 
The learning goal is to minimize the difference between the input $\mathbf{x}$ and the reconstructed $\hat{\mathbf{x}}$.





    


Given a set of training inputs $\mathbf{X} = [\mathbf{x}_{1}, \mathbf{x}_{2}, \dots, \mathbf{x}_{C}]$, the corresponding autoencoder outputs $\mathbf{\hat{X}} = [\mathbf{\hat{x}}_{1}, \mathbf{\hat{x}}_{2}, \dots, \mathbf{\hat{x}}_{C}]$, where $\mathbf{\hat{x}}_{i}=\mathrm{AE}(\mathbf{x}_{i})$, and the learnable parameters $\theta_{AE}$ of the \texttt{AE}, 
the objective function $\mathcal{L}_{AE}$ used for training is formulated in Equation~\ref{eqn:objective_ae}. 
\begin{align}
    \arg\min_{\theta_{AE}} \mathcal{L}_{AE} &= \arg\min_{\theta_{AE}} \frac{1}{C}\sum^{C}_{i=1} (\mathbf{x}_{i} - \mathbf{\hat{x}}_i)^{2}
    \label{eqn:objective_ae}
\end{align}
The reconstruction error (\textit{RE}) is defined as 
$||\mathbf{x}_{i}-\hat{\mathbf{x}}_{i}||^2_2$. 
If the \textit{RE} exceeds threshold $\epsilon$, 
$\mathbf{x}_{i}$ is considered as an outlier~\cite{Sakurada14}.

Applications of \texttt{AE}s often relate to non-sequential data, such as images, and \texttt{AE}s 
do not consider relationships between observations over time, e.g., the increasing trend as shown in Figure~\ref{fig:reconstruction}, as summarized in Table~\ref{table:summary}.

\begin{table}[ht!]
\centering
\small
\caption{Autoencoder-based models for outlier detection.}
\label{table:summary}
\begin{tabular}{c|P{1.4cm}|c|c|}
\cline{2-4}
& \textit{Temporal-dependencies} 
& \textit{Efficiency} & \textit{Diversity} \\ \hline
\multicolumn{1}{|l|}{\texttt{AE}} & \xmark  & \cmark & \xmark \\ \hline
\multicolumn{1}{|l|}{\texttt{RAE}} & \cmark & \xmark & \xmark \\ \hline
\multicolumn{1}{|l|}{\texttt{CAE}} & \cmark & \cmark & \xmark \\ \hline
\multicolumn{1}{|l|}{\texttt{AE-Ensemble}} & \xmark  & \cmark & \cmark~(Implicit) \\ \hline
\multicolumn{1}{|l|}{\texttt{RAE-Ensemble}} & \cmark  & \xmark & \cmark~(Implicit) \\ \hline
\multicolumn{1}{|p{1.8cm}|}{\texttt{CAE-Ensemble}} & \cmark & \cmark & \cmark~(Explicit) \\ \hline
\end{tabular}
\end{table}

\noindent
\textbf{Recurrent Autoencoders. }
%
A sequence-to-sequence model targets the analysis of sequential data, 
by preserving the relationships surrounding each data point, e.g., temporal dependencies in time series.  
For example, if a time series is growing, e.g., as shown in Figure~\ref{fig:reconstruction}, such a model is capable to 
taking this growing tendency into account when making predictions or detecting outliers. 

%
A recurrent autoencoder (\texttt{RAE}) follows a sequence-to-sequence model using a recurrent neural network (\texttt{RNN}). 
In the encoder, an observation $\mathbf{s}_{t}$ in time series $\mathcal{T}$ and a hidden state at the previous timestamp $\mathbf{h}^{(E)}_{t-1}$ are fed into an \texttt{RNN} unit 
to compute the current hidden state 
$\mathbf{h}^{(E)}_{t}$ at timestamp $t$ using 
Equation~\ref{eq:rnn}. 
%
\begin{equation}
\mathbf{h}^{(E)}_{t} = \mathrm{RNN}(\mathbf{s}_{t}, \mathbf{h}^{(E)}_{t-1}).  \label{eq:rnn}
\end{equation}
Here, $\mathrm{RNN}(\cdot)$ is an abstraction that can be 
a Long Short Term Memory (\texttt{LSTM})~\cite{Hochreiter97}
or a Gated Recurrent Unit (\texttt{GRU})~\cite{Dauphin17}. Once $\mathbf{h}^{(E)}_{t}$ is computed, it is fed into the next \texttt{RNN} unit to compute the hidden state at timestamp $t+1$.

In the decoder, the time series is reconstructed in reverse order. Specifically, the last hidden state in the encoder $\mathbf{h}^{(E)}_{C}$ is used as the first hidden state of the decoder $\mathbf{h}^{(D)}_{C}$, i.e., $\mathbf{h}^{(E)}_{C} = \mathbf{h}^{(D)}_{C}$. Based on the previous hidden state $\mathbf{h}^{(D)}_{t+1}$
of the decoder and the previous reconstructed observation $\mathbf{\hat{s}}_{t+1}$, the decoder reconstructs the current observation $\mathbf{\hat{s}}_{t} = \mathrm{RNN}(\mathbf{\hat{s}}_{t+1}, \mathbf{h}^{(D)}_{t+1})$ using an \texttt{RNN} unit. 


Similar to an \texttt{AE}, an \texttt{RAE} identifies the learnable parameters 
that minimize the difference between the reconstructed time series and the original time series. 
%
%
Then, given the reconstructed observations represented by $\mathcal{\hat{T}} = \langle \hat{\mathbf{s}_{1}}, \hat{\mathbf{s}_{2}}, \dots, \hat{\mathbf{s}_{C}} \rangle$, it is possible to calculate the differences from the original time series $\mathcal{T}$. 
Specifically, the outlier score for observation $\mathbf{s}_{i}$ is defined as  $||\mathbf{s}_{i} - \hat{\mathbf{s}_{i}}||^{2}_{2}$. 

Despite the good accuracy achieved by \texttt{RAE}~\cite{Kieu19}, they suffer from low efficiency. 
In an \texttt{RNN}, the computation of one state takes as input the previous state, 
making \texttt{RNN} learning unparallelizable~\cite{Malhotra16}. 
To improve efficiency, we propose a convolutional sequence-to-sequence autoencoder \texttt{CAE} in Section~\ref{subs:ConvSeq} that captures temporal dependencies without recursive computations.  

\noindent
\textbf{Autoencoder Ensembles. }
Ensemble models combine multiple, diverse basic models and often achieve 
better accuracy than a single model by avoiding overfiting and underfitting (a.k.a., achieving better bias-variance trade-off) . 
%
%
%
To achieve improved accuracy, ensembles need to be designed so that the individual basic models are diverse and contribute differently to solving the  problem~\cite{Aggarwal17}. %

Existing autoencoder ensembles often create different basic models randomly, without explicitly quantifying their diversity.
For example, \texttt{AE-Ensemble}~\cite{Chen17} creates multiple \texttt{AE}s by removing connections between neurons at random in   feed-forward neural networks, and
\texttt{RAE-Ensemble}~\cite{Kieu19} incorporates random skip connections to generate \texttt{RAE}s with different network structures. 
In other words, existing autoencoder ensembles consider diversity only implicitly. %
%
In Section~\ref{subs:Diversity}, we propose a metric to  quantify the diversities among different basic models and incorporate this metric explicitly into the objective function, making diversity part of the learning goals. 
This explains the ``Diversity'' column in Table~\ref{table:summary}. 

\noindent
\textbf{Summary. }
Table~\ref{table:summary} summarizes the existing autoencoder based models. 
The first three do not use ensembles and thus do not offer any diversity. \texttt{AE} is efficient, as matrix computations can be parallelized easily, but is unable to capture temporal dependencies. 
\texttt{RAE} is able to capture temporal dependencies, but is inefficient due to its recursive computations. We propose an efficient \texttt{CAE} that is able to capture temporal dependencies. We offer empirical evidence to justify this in Section~\ref{experiments}.

The last three models are ensembles whose basic models employ the first three models, respectively. 
Thus, they have the same properties as their corresponding basic models. 
%
%
\texttt{AE-Ensemble} and \texttt{RAE-Ensemble} offer limited diversity because they generate basic models with randomly selected network structures without explicitly quantify the diversity in the learning processes. In contrast, we propose a metric that quantifies the diversity between basic models, and we judiciously design an objective function to incorporate the diversity metric to make \texttt{CAE-Ensemble} a diversity-driven ensemble. In addition, we propose a parameter sharing mechanism to improve the training efficiency of \texttt{CAE-Ensemble}. 



\section{The \texttt{CAE-Ensemble} Framework}
\label{framework}

\begin{figure}[!htb]
    \centering
    \includegraphics[width=0.95\linewidth]{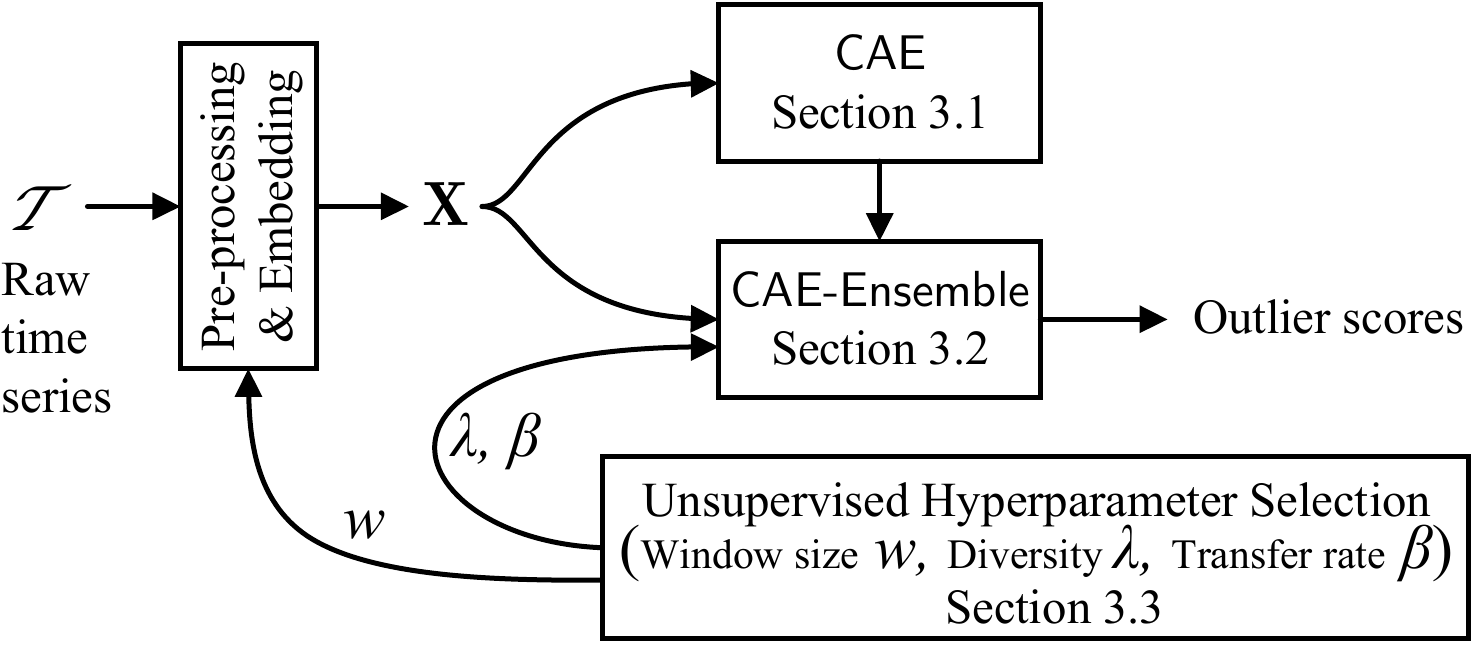}
    \caption{Framework overview.}
    \label{fig:framework_overview}
\end{figure}

This section details the proposed \texttt{CAE-Ensemble}. In Section~\ref{subs:ConvSeq}, we first cover basic \texttt{CAE} models using convolutional sequence-to-sequence autoencoders that are able to capture temporal dependencies in time series data while ensuring efficiency. Next, in Section~\ref{subs:Diversity}, we describe a diversity-driven ensemble \texttt{CAE-Ensemble} that consists of the basic \texttt{CAE} models while taking into account diversity during training, along with an efficient training method using parameter transferring for the proposed ensemble. Finally, in Section~\ref{parameters_choice}, we propose an strategy for tuning hyperparameters in a fully unsupervised manner. 
The complete process is summarized in Algorithm~1 and a framework overview is shown 
in Figure~\ref{fig:framework_overview}. 
\begin{algorithm}[!htp]
    \small
    \SetKwFunction{Normalization}{$\mathit{ReScale}$}
    \SetKwFunction{SlidingWindows}{$\mathit{SplitIntoWindows}$}
    \SetKwFunction{Hyper}{$\mathit{HyperparameterSelection}$}
    \SetKwFunction{Embedding}{$\mathit{Embedding}$}
    \SetKwFunction{Optimize}{$\mathit{OptimizeNormal}$}
    \SetKwFunction{OptimizeD}{$\mathit{OptimizeDiverse}$}
    \SetKwInOut{KwIn}{Input}
    \SetKwInOut{KwOut}{Output}

    \KwIn{Raw time series $\mathcal{T}$, Number of Basic Models $M$}
    \KwOut{Outlier scores}
    $\mathit{outlierScores}[] \leftarrow \emptyset, \,\mathit{savedParam} \leftarrow \emptyset$;
    
    $\mathcal{T} \leftarrow \Normalization(\mathcal{T})$; \\
    $\beta, \lambda, w \leftarrow \Hyper(\mathcal{T})$; \tcc{Sec 3.3}  \label{model:line:hyper}
    
    $\mathcal{T}_{\mathit{windows}} \leftarrow \SlidingWindows(\mathcal{T},w)$;

    $\mathbf{X} \leftarrow \Embedding(\mathcal{T}_{\mathit{windows}})$;
    

    \For{$i \leftarrow 1$ \KwTo $M$}{ \label{model:line:ensemStart}
        \eIf{$i$ = 1}{

            
            \tcc{Build the first basic model CAE $f_1$.} 
            
            $\theta_{f_1} \leftarrow$ \Optimize ($f_1$); \\
             $\mathbf{\hat{X}^{(1)}} \leftarrow f_{1}(\mathbf{X})$;
         }{
            \tcc{Build the $i$-th basic model CAE $f_i$.} 
            
            $\theta_{f_i} \leftarrow \OptimizeD(f_i,\lambda, \mathit{savedParam})$; 
            \tcc{\OptimizeD uses Eq.~\ref{eqn:objective}.
            }
            
            $\mathbf{\hat{X}^{(i)}} \leftarrow f_{i}(\mathbf{X})$; 
            
         }
        
        
        $\mathit{outlierScores}[i] \leftarrow
        \langle \|\mathbf{x}_{1}-\hat{\mathbf{x}}_{1}^{(i)}\|_2^2, \ldots, \|\mathbf{x}_{w}-\hat{\mathbf{x}}_{w}^{(i)}\|_2^2 \rangle$;
        
        $\mathit{savedParam} \leftarrow$ Randomly select the fraction $\beta$ of the parameters $\theta_{f_i}$ from basic model CAE $f_i$;
        
    } 
    \KwRet{$\mathit{median(outlierScores)}$} 
    \caption{CAE-Ensemble} 
\end{algorithm}

Before introducing the model, we cover a pre-processing step: 
a \textit{raw time series} is pre-processed into \textit{time series windows} that are then used for training and testing~\cite{DBLP:conf/cikm/CirsteaMMG018}.  
The pre-processing first 
re-scales an observation $x$ in the time series to 
%
 $\displaystyle z =\frac{x-\mu}{\sigma}$, 
%
where $\mu$ is the mean and $\sigma$ is the standard deviation of the observations in the training time series. 
%
%
This is to prevent that magnitude differences among different dimensions in a time series affect the reconstruction errors differently, which is a common pre-processing technique~\cite{Sammut17}.


%
We then create sliding windows of size $w$ that slide one observation at a time. 
For example, for $\mathcal{T} = \langle \mathbf{s}_{1}, \mathbf{s}_{2}, \dots, \mathbf{s}_{C} \rangle$, the first window is $\langle \mathbf{s}_{1}, \mathbf{s}_{2}, \dots, \mathbf{s}_{w} \rangle$ and the second is $\langle \mathbf{s}_{2}, \mathbf{s}_{3}, \dots, \mathbf{s}_{w+1} \rangle$, etc.  
%


%

%
%



\subsection{Convolutional 
Autoencoder \texttt{CAE}}\label{subs:ConvSeq}
\begin{figure}[!tbh]
    \centering
    \includegraphics[width=0.9\linewidth]{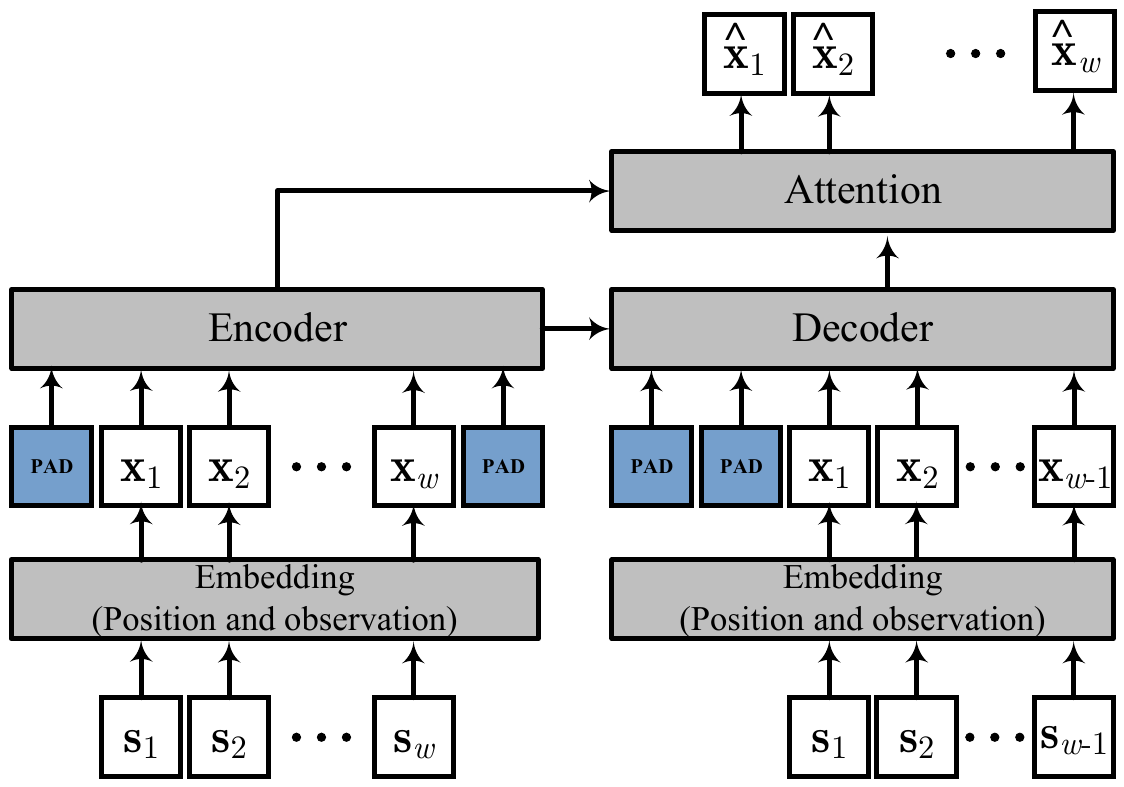}
    \caption{Convolutional  autoencoder \texttt{CAE}.
    }
    \label{fig:cnn_seqtoseq}
\end{figure}

We build the \texttt{CAE-Ensemble} from basic models that use convolution sequence-to-sequence autoencoders \texttt{CAE}s. 
%
A \texttt{CAE} combines convolutional neural networks (\texttt{CNN}s) with a sequence-to-sequence architecture, as shown in Figure~\ref{fig:cnn_seqtoseq}.   
%
First, the \texttt{CAE} embeds a time series window $\mathcal{T}$ into vectors that capture both the content and positions of the observations in $\mathcal{T}$. 
These vectors are then fed to the encoder that employs a 1D \texttt{CNN} to extract features that capture temporal dependencies in $\mathcal{T}$ and then outputs the features as hidden states. 
Next, the decoder employs another 1D \texttt{CNN} to extract features from both the embedded vectors and the hidden states from the encoder. 
The output from the decoder is another set of hidden states. 
As a 1D \texttt{CNN} does not involve recursive computations, it is possible to perform 1D \texttt{CNN} for different timestamps in parallel. This improves efficiency.
Finally, an attention layer is applied to combine the hidden states from the encoder and the decoder, the result of which is then used for reconstructing the time series. 
%
%
%


\subsubsection{Embedding}
The embedding encompasses: \textit{observation embedding} and \textit{position embedding}.

\noindent\textbf{Observation Embedding:}
The input of the encoder is a time series window  
$\mathcal{T} = \langle \mathbf{s}_{1}, \mathbf{s}_{2}, \dots, \mathbf{s}_{w} \rangle$, with vectors $\mathbf{s}_{t} \in \mathbb{R}^{D}$, which is then embedded into $\mathbf{V} = \langle \mathbf{v}_{1}, \mathbf{v}_{2}, \dots, \mathbf{v}_{w} \rangle$, where each vector $\mathbf{v}_{t} \in \mathbb{R}^{D'}$ is a $D'$ dimensional representation for the original $D$ dimensional observation $\mathbf{s}_{t}$ 
that captures its most typical features~\cite{Hu16}. 

Specifically, we have $\mathbf{v}_{t} = f_s(\mathbf{W}_{v} \cdot \mathbf{s}_{t} + \mathbf{b}_{v})$, 
where $f_s$ represents a non-linear activation function that maps the original time series observation $\mathbf{s}_{t}$ to the embedding and $\mathbf{W}_{v} \in \mathbb{R}^{D' \times D}$ and $\mathbf{b}_{v} \in \mathbb{R}^{D'}$ are learnable parameters.
%


\noindent\textbf{Position Embedding:} 
When using \texttt{CNN}s instead of \texttt{RNN}s to model time series, we need to capture explicitly the positions of the observations in time series. 
%
%
Thus, we define a positional entry $t$ for each observation in a time series. 
For example, the 5-th observation in a time series has the position entry 5.
Specifically, positional entries $1, 2, \dots, w$ are embedded into $\mathbf{p}_{1}, \mathbf{p}_{2}, \dots, \mathbf{p}_{w}$, respectively, where each $\mathbf{p}_{t} \in \mathbb{R}^{D'}$, with $t\in[1, w]$, is a $D'$ dimensional representation of the position of an observation. 
The transformation $\mathbf{p}_{t} = f_t(\mathbf{W}_{p} \cdot t + \mathbf{b}_{p})$ 
is similar to the observation embedding. 
%

\begin{figure}[!ht]
    \centering
    \includegraphics[width=0.9\linewidth]{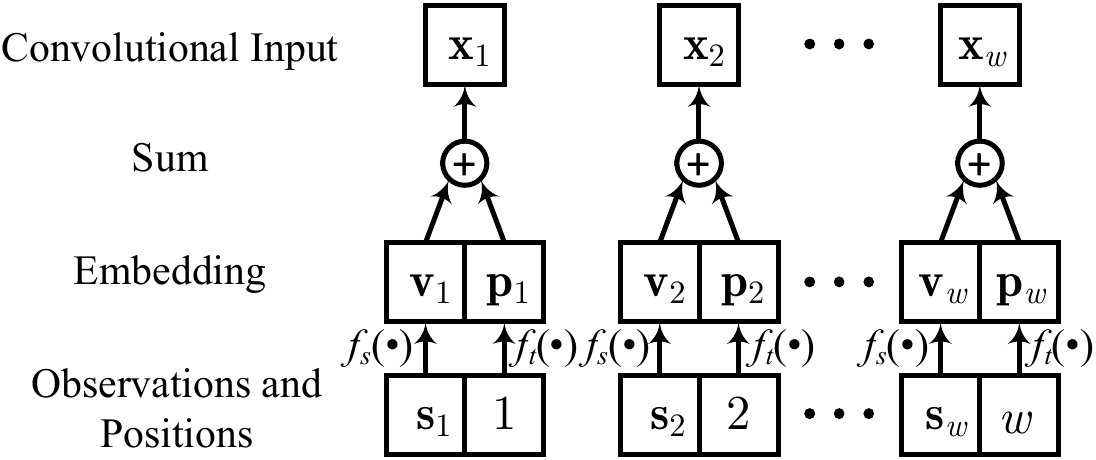}
    \caption{Embedding process.
    }
    \label{fig:positionalInput}
\end{figure}

\noindent\textbf{Final Embedding:}
The observation and position embeddings are combined to obtain $\mathbf{X} = \langle \mathbf{x}_{1}, \mathbf{x}_{2}, \dots, \mathbf{x}_{w} \rangle = \langle \mathbf{v}_{1} + \mathbf{p}_{1}, \mathbf{v}_{2} + \mathbf{p}_{2}, \dots, \mathbf{v}_{w} + \mathbf{p}_{w} \rangle$ as the input to the \texttt{CNN}. 
We use sum, instead of, e.g., concatenation, as existing studies show that summing achieves high effectiveness~\cite{Gehring17,Vaswani17}. 
%
A summary of the process is shown in Figure~\ref{fig:positionalInput}.

\subsubsection{Encoder}
\begin{figure}[ht!]
    \centering
    \includegraphics[width=0.8\linewidth]{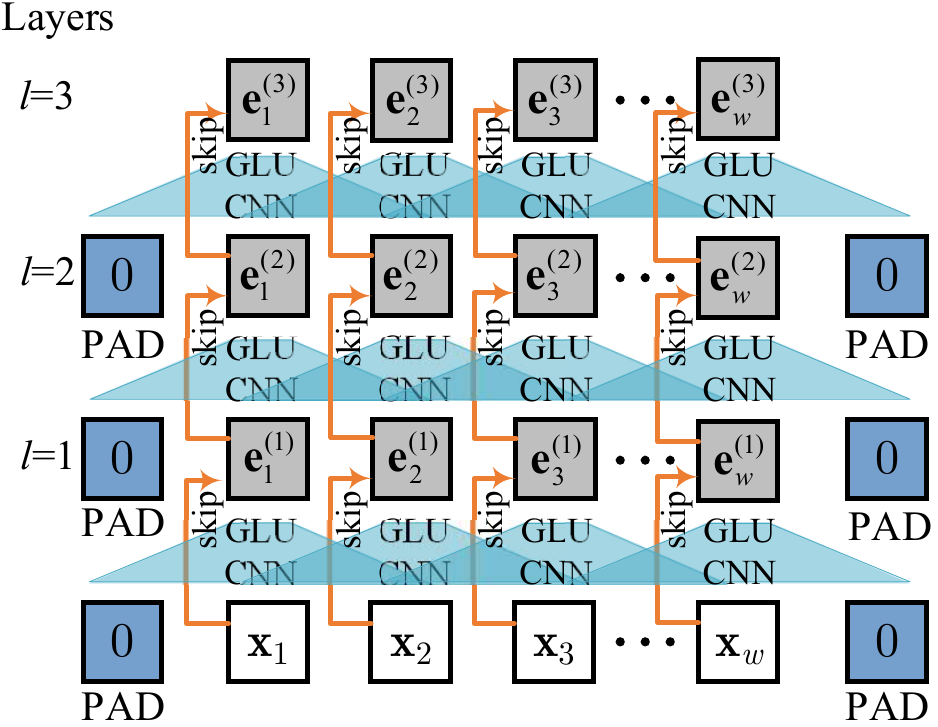}
    \caption{An example encoder with 3 layers.}
    \label{fig:encoder}
\end{figure}

The encoder employs a stack of convolutional layers to capture typical temporal features and trends of time series.  Figure~\ref{fig:encoder} shows an encoder with 3 convolutional layers. 
Each layer is framed by a 1D convolution followed by a non-linear activation function. 
More specifically, the $l$-th layer takes as input a vector $\mathbf{E}^{(l)} \in \mathbb{R}^{w \times D'}$ and feeds a new vector $\mathbf{E}^{(l+1)} \in \mathbb{R}^{w \times D'}$ into the next layer, i.e., $(l+1)$-th convolution layer. 
The $\mathbf{E}$ vectors are the hidden states of the encoder.

\begin{align}
   \mathbf{E}^{(l+1)} &= f_E(\mathbf{W}_{E}^{(l)} \otimes \mathrm{GLU}(\mathbf{E}^{(l)}) + \mathbf{b}_{E}^{(l)})
\end{align}

The input to the first layer, $\mathbf{E}^{(0)}$, is the embedded input vector $\mathbf{X}$ with zero padding. We use padding  in all layers to ensure that the output of the convolution has the same length as the input. 
%
%
Next, $\otimes$ denotes the 1D convolutional operator; $k$ is the kernel size of the convolution operator; and 
$\mathbf{W}_{E}^{(l)} \in \mathbb{R}^{D' \times k}$ 
and  $\mathbf{b}_{E}^{(l)} \in \mathbb{R}^{D'}$ are the kernel matrix and bias vector, respectively, at the $l$-th layer. 
Multiple different kernel matrices are often used. 
$f_E(\cdot)$ is a non-linear activation function.

To capture temporal information flow in time series better, we integrate Gated Linear Units (\texttt{GLU}s)~\cite{Dauphin17} into the convolution layers. Specifically, a \texttt{GLU} is applied to $\mathbf{E}^{(l)}$ before applying the convolution operation. 
A \texttt{GLU} mimics the gating mechanisms used in \texttt{RNN}s and controls how much information along the temporal dimension should be kept or forgotten. 
First, from $\mathbf{E}^{(l)}$, we produce $\mathbf{A}^{(l)}_{1}$ and $\mathbf{A}^{(l)}_{2}$ by applying convolution (cf. Equation~\ref{eq:A}).  
Here, 
$\mathbf{W}^{(l)}_{A_{i}} \in \mathbb{R}^{D' \times k}$ and $\mathbf{b}_{A_{i}} \in \mathbb{R}^{D'}$ are the kernel matrix and bias vector, respectively. 
%
Then, we perform element-wise matrix multiplication, denoted by $\odot$,  between  $\mathbf{A}^{(l)}_{1}$ and the $\mathrm{sigmoid}$ function applied to $\mathbf{A}^{(l)}_{2}$, thus controlling how much information should be kept or forgotten (see Equation~\ref{eq:GLU}). 
%
%
\texttt{GLU} is specified in Equations~\ref{eq:GLU} and \ref{eq:A}. 

\begin{align}
  \mathrm{GLU}(\mathbf{E}^{(l)}) = \mathbf{A}^{(l)}_{1} \odot \sigma(\mathbf{A}^{(l)}_{2}) \label{eq:GLU}
  \end{align}
  \begin{align}
  \mathbf{A}^{(l)}_{1} = \mathbf{W}^{(l)}_{A_{1}} \otimes \mathbf{E}^{(l)} + \mathbf{b}_{A_{1}}, ~~~
  \mathbf{A}^{(l)}_{2} = \mathbf{W}^{(l)}_{A_{2}} \otimes \mathbf{E}^{(l)} + \mathbf{b}_{A_{2}} 
  \label{eq:A}
\end{align}


Finally, at each convolution layer, the output of the current convolution layer is added to the output from the previous convolution layer by using a skip or residual connection. Formally, we have $\mathbf{E}^{(l+1)} = \mathbf{E}^{(l+1)} + \mathbf{E}^{(l)}$. 
%
%
Using skip connections establish a flow for gradients across layers in the network, 
reducing the vanishing and the exploding gradients issues~\cite{He16}. 
%

%


The encoder process is summarized in Figure~\ref{fig:encoder}, where the blue triangles represent the 1D convolution with \texttt{GLU} and the arrows indicate the skip connections.

\subsubsection{Decoder}
\begin{figure}[t]
    \centering
    \includegraphics[width=0.8\linewidth]{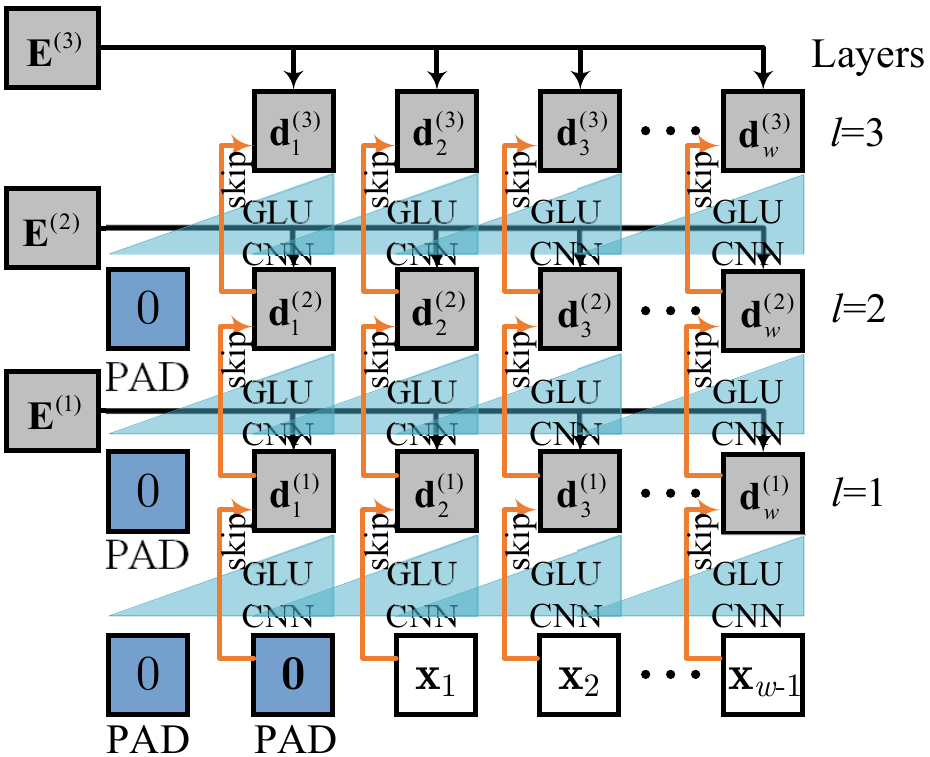}
    \caption{An example decoder with 3 layers. 
    }
    \label{fig:decoder}
\end{figure}

The decoder shown in Figure~\ref{fig:decoder} is slightly different from the encoder. In the decoder, 
%
we ensure that the convolution at timestamp $t$ only uses observations that appear no later than $t$. This aligns with the time series setting, where observations only to be seen in the future cannot be utilized. 
Thus, we apply padding to the input vector only before the first observation and then  
compute the hidden states for the decoder $\mathbf{D} \in \mathbb{R}^{w \times D'}$ as follows.

\begin{align}
   \mathbf{D}^{(l+1)} &= f_D(\mathbf{W}_{D}^{(l)} \otimes \mathrm{GLU}(\mathbf{D}^{(l)}) + \mathbf{b}_{D}^{(l)} + \mathbf{E}^{(l)})
\end{align}
Here, $\otimes$ denotes the 1D convolution; $k$ is the kernel size of the convolution operator; $\mathbf{W}_{D}^{(l)}$ and $\mathbf{b}_{D}^{(l)}$ are the kernel matrix and the bias vector at the $l$-th layer, respectively; $\mathbf{E}^{(l)}$ is the hidden state from the encoder at the same layer; and $f_D(\cdot)$ is a non-linear activation function.
%
%

As for the encoder, at each layer, the output of the current layer is added to the output from the previous layer by using a skip connection, i.e., $\mathbf{D}^{(l+1)} = \mathbf{D}^{(l+1)} + \mathbf{D}^{(l)}$.

\subsubsection{Attention}

We apply an attention mechanism~\cite{Luong15} in the convolutional sequence-to-sequence autoencoder to capture local temporal patterns, e.g., periodicity, in an input time series by identifying similar observations that reoccur. 
%
Thus, the attention should consider observations that are relatively more important than others, given different specific settings. 

To achieve this, we adopt a global attention method~\cite{Luong15} that 
computes a context vector {$\mathbf{c}^{(l)}_{t}$ that captures the relative importance of the observation} at each timestamp $t$ {for each decoder layer $l$}. 
%
Specifically, 
we first compute a state summary 
%
$\mathbf{z}^{(l)}_{t} = \mathbf{W}^{(l)}_{z} \cdot \mathbf{d}^{(l)}_{t} + \mathbf{b}^{(l)}_{z}$ 
using the decoder hidden state $\mathbf{d}^{(l)}_{t}$ at each timestamp. 
%
%
Then, we compute an attention score $\alpha^{(l)}_{tt'}$ that captures the temporal correlation between the observation at timestamp $t$ at decoder layer $l$ and the observation at timestamp $t'$ at encoder layer $l$.
Attention score $\alpha^{(l)}_{tt'}$ is computed 
as follows.
\begin{align}
    \alpha^{(l)}_{tt'} = \frac{\mathrm{exp}(\mathbf{z}^{(l)}_{t} \cdot \mathbf{e}^{(l)}_{t'})}{\sum_{t'=1}^{w}\mathrm{exp}(\mathbf{z}^{(l)}_{t} \cdot \mathbf{e}^{(l)}_{t'})}
\end{align}
Here, $\mathbf{e}^{(l)}_{t'}$, $t'\in[1,w]$, is the output of encoder layer $l$ at timestamp $t$, and $\cdot$ denotes the dot product.

\begin{figure}[ht!]
    \centering
    \includegraphics[width=0.80\linewidth]{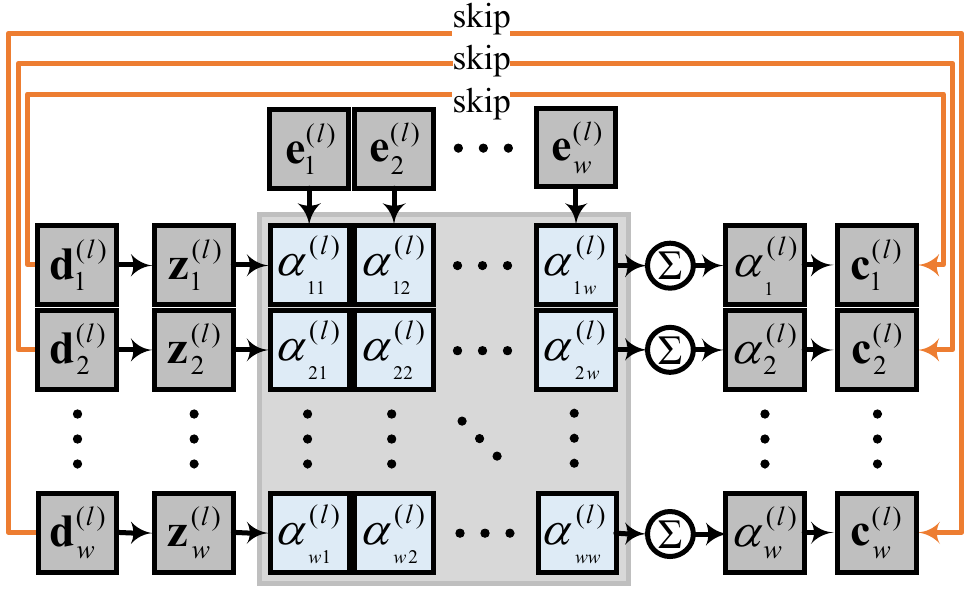}
    \caption{Attention structure.}
    \label{fig:attention}
\end{figure}

Next, a context vector 
$\mathbf{c}^{(l)}_{t} = \sum^{w}_{t'=1} \alpha^{(l)}_{tt'} \mathbf{e}^{(l)}_{t'}$
captures the temporal correlation between the observation at timestamp $t$ at the decoder layer $l$ and all observations at the encoder layer $l$. 
%
%
%
Once $\mathbf{c}^{(l)}_{t}$ is computed, we update the output of the corresponding decoder layer $\mathbf{d}^{(l)}_{t}$ by replacing $\mathbf{d}^{(l)}_{t}$ by $\mathbf{c}^{(l)}_{t} + \mathbf{d}^{(l)}_{t}$, i.e., $\mathbf{D}^{(l)} = \mathbf{C}^{(l)} + \mathbf{D}^{(l)}$.
In Figure~\ref{fig:attention}, we summarize the attention mechanism. 
The attention score $\alpha^{(l)}_{t}$ between the decoder state summary and the encoder hidden states is calculated and weighted with the embedding information to obtain the context vector $\mathbf{c}^{(l)}_{t}$. Finally, represented by the orange lines, the output for the decoder is updated using the context to obtain the reconstruction $\hat{\mathbf{X}}$.

\subsubsection{Reconstruction}
At the last $L$-th convolution layer in the decoder, we obtain a final hidden state $\mathbf{D}^{(L+1)}$. We reconstruct the time series from this state using a simple fully connected neural network. Specifically, we have $\hat{\mathbf{X}} = f_{R}(\mathbf{W}_{R} \otimes \mathrm{GLU}(\mathbf{D}^{(L+1)}) + \mathbf{b}_{R})$. 
%
Here, $\hat{\mathbf{X}} = \langle \hat{\mathbf{x}_{1}}, \hat{\mathbf{x}_{2}}, \dots, \hat{\mathbf{x}_{w}} \rangle$ is the reconstruction of time series $\mathbf{X}$; $\mathbf{W}_{R}$ and $\mathbf{b}_{R}$ are the weight matrix and bias vector of the reconstruction layer, respectively; function $f_R(\cdot)$ is a non-linear activation function.
The reconstruction errors between  the elements of $\hat{\mathbf{X}}$ and $\mathbf{X}$ are used as outlier scores.




\subsection{Diversity-Driven Ensembles}\label{subs:Diversity}

We use the proposed \texttt{CAE}s as basic models in the ensemble called \texttt{CAE-Ensemble}. 
Ensembles are generally able to improve overall accuracy by combining the outputs from individual basic models~\cite{Chen17}. 
%
A naive approach is to first train multiple basic models. Then, each basic model is used to reconstruct embedded time series $\mathbf{X}$. The average over the reconstructed time series from all basic models is used as the final reconstructed time series for the ensemble, as defined in Equation~\ref{eq:final_reconstruction}.
\begin{align}
   F(\mathbf{X}) = \frac{1}{M} \sum_{m=1}^{M} f_{m}(\mathbf{X}),
   \label{eq:final_reconstruction}
\end{align}
where $M$ is the total number of basic models, $f_{m}(\cdot)$ refers to the $m$-th basic model, and $F(\cdot)$ is the output of the ensemble. 
%


This naive approach has two limitations: (1) If the basic models are similar, the accuracy of the ensemble is similar to those of the basic models, meaning that the ensemble is not substantially better than each basic model. 
%
(2) The computational cost of the ensemble is often $M$ times that of a single basic model. When $M$ is large, the ensemble can be very expensive to train. 

The first limitation has been addressed partially by using basic models with different structures, e.g., by randomly modifying the connections among computational units~\cite{Chen17, Kieu19}. 
However, basic models with different structures may not produce diverse outputs. 
%
%
%
%
%
%

\begin{figure}[ht!]
    \centering
    \includegraphics[width=1\linewidth]{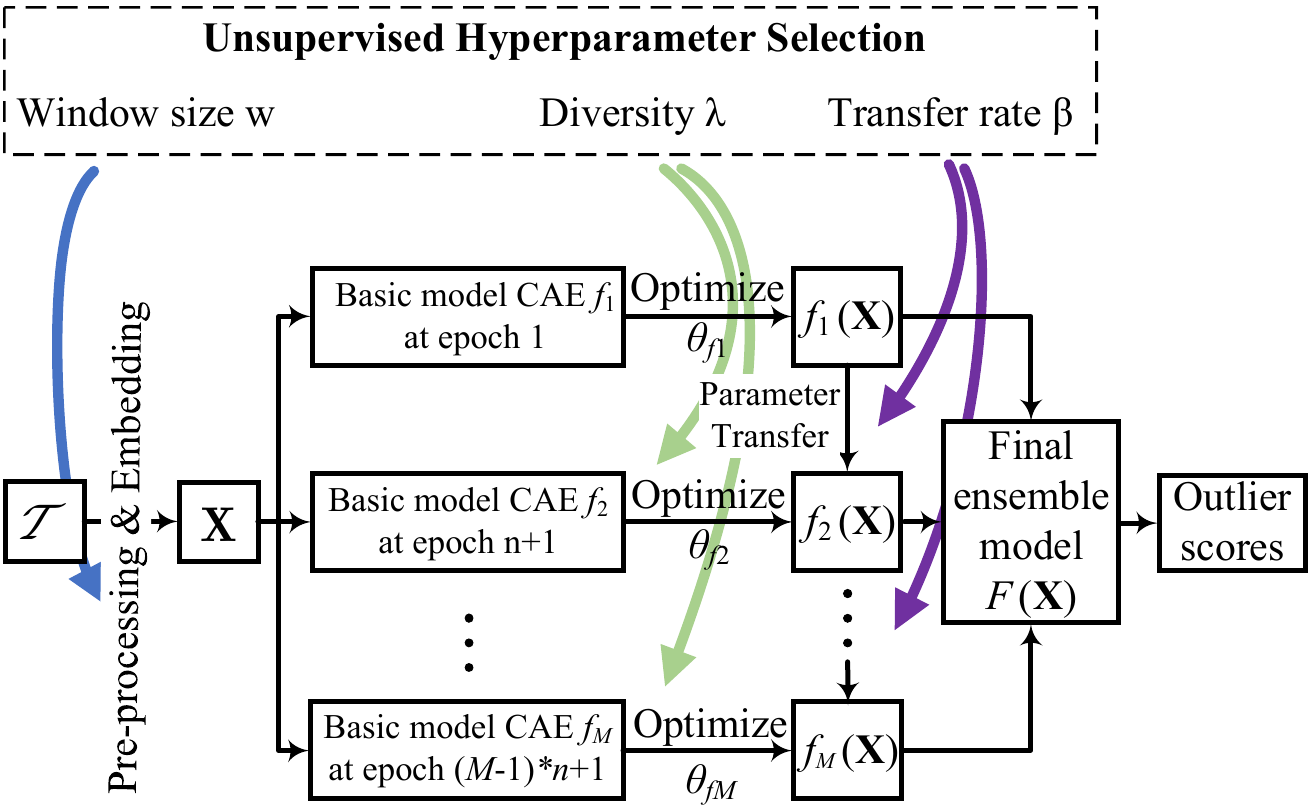}
    \caption{\texttt{CAE-Ensemble} overview.}
    \label{fig:diversityOptimization}
\end{figure}

We introduce a diversity-driven ensemble to address the above two limitations, as depicted in  
Figure~\ref{fig:diversityOptimization}. 
Rather than training different basic models independently~\cite{Chen17, Kieu19}, we generate the basic models one by one. 
When training a basic model, we design an objective function that considers not only the accuracy of the model but also its difference compared to the previous basic models, which ensures diversity among basic models, thus addressing the first limitation.  
In addition, when training a basic model, we transfer a portion of the parameters from the previous basic model to the model instead of training the model from scratch. This helps significantly reduce the training time, thus addressing the second limitation. 

\subsubsection{Basic Model Generation}

%
Inspired by Born-again Neural Networks~\cite{Furlanello18} (\texttt{BANN}) and AdaBoost Negative Correlation~\cite{Wang10}, 
we iteratively generate basic models in the model training epochs. 
For example, we may generate a basic model per $n$ training epochs. If $n=10$, when training an ensemble using 200 epochs, we can then generate 20 basic models. 

In the first epoch, we create the first basic model $f_1(\cdot)$, and we start training this basic model using embedded time series $\mathbf{X}$. We then add the first basic model to the ensemble. After $n$ epochs, the first basic model is partially trained. 
We continue to generate the second basic model $f_2(\cdot)$ by transferring a portion of the parameters learned in the first basic model to $f_2(\cdot)$. Then, we train the second basic model. 
The training and basic model generation processes continue until the last training epoch. 

%

\begin{figure}[ht!]
    \centering
    \includegraphics[width=0.65\linewidth]{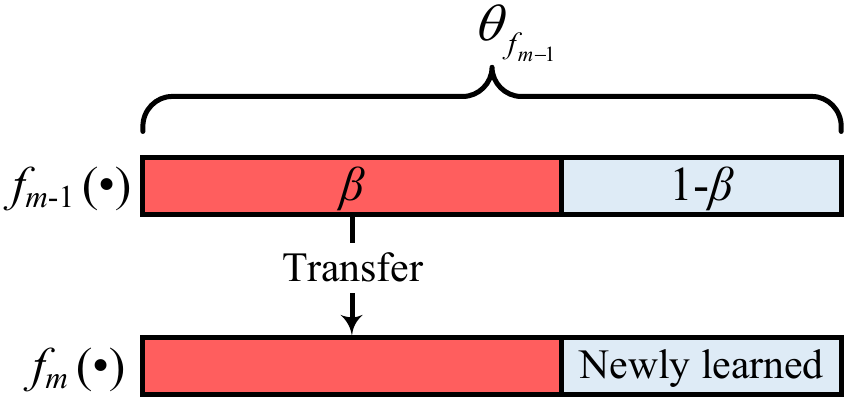}
    \caption{Basic model generation with parameter transfer.
    }
    \label{fig:transfer}
\end{figure}

More specifically, given a basic model $f_{m-1}(\cdot)$ with $\theta_{f_{m-1}}$ trained parameters obtained in the previous epochs, the newly generated basic model $f_{m}(\cdot)$ receives a randomly selected fraction $\beta$ of its parameters from $f_{m-1}(\cdot)$. Then, the remaining fraction $1-\beta$ of the parameters must be trained in subsequent epochs. 
This approach enables us to train a large number of ensemble components with low training time. 
Figure~\ref{fig:transfer} shows the process of generating new basic models with parameter transfer. 
%
%

Note that the proposed ensemble differs from a Snapshot Ensemble~\cite{Huang17}, where the trained parameters are transferred completely among the basic models.
%

\subsubsection{Diversity Metric} Based on existing studies~\cite{Aggarwal17}, an ensemble benefits from consisting of diverse basic models. 
The more diverse the basic models in an ensemble are, the more accurate the ensemble often achieves. 
%
To quantify the diversity explicitly, we define a diversity metric 
$\mathit{DIV}_{f_{m}, f_{n}}(\cdot)$ to measure the dissimilarity 
between two basic models $f_{m}(\cdot)$ and $f_{n}(\cdot)$. 
%
\begin{align}\label{eq:diversity}
   \mathit{DIV}_{f_{m}, f_{n}}(\mathbf{X}) = 
   {||f_{m}(\mathbf{X}) - f_{n}(\mathbf{X})||_{2}},
\end{align}
Here, a larger $\mathit{DIV}_{f_{m}, f_{n}}(\mathbf{X})$ value indicates a larger difference between the outputs of basic models $f_{m}(\cdot)$ and $f_{n}(\cdot)$.

The design of diversity metric $\mathit{DIV}_{f_{m}, f_{n}}(\cdot)$ is inspired by the supervised diversity metric~\cite{Zhang20}. However, the supervised diversity metric computes the $\mathrm{softmax}$ outputs of two basic models and compares the outputs with ground-truth labels to quantify the diversity. 
In our unsupervised setting, we do not have ground truth labels. Instead, we aim for different basic models with different reconstructions, i.e., basic models with diverse outputs.  

Next, we extend the diversity metric $\mathit{DIV}_{F}(\cdot)$ to measure the diversity of an ensemble model $F(\cdot)$ as follows.
\begin{align}\label{eq:divensem}
   \mathit{DIV}_{F}(\mathbf{X}) = \frac{2}{M(M-1)}\sum^{M}_{m=1}\sum^{M}_{n=m+1}\mathit{DIV}_{f_{m}, f_{n}}(\mathbf{X}),
\end{align}
where a large $\mathit{DIV}_{F}(\cdot)$ value indicates that ensemble model $F(\cdot)$ is more diverse. 

\subsubsection{Diversity-Driven Objective Function} 
Based on the proposed diversity metric, we define the objective function for training each basic model in the ensemble in two parts. 

First, a basic model $f_{m}(\cdot)$ should reconstruct the time series accurately, which is the same as for all autoencoder models. Thus, the objective function $\mathcal{J}_{f_{m}}$ measures the difference between the input embedded time series vectors
and the reconstructed vectors.
\begin{align}
    \mathcal{J}_{f_{m}} = ||\mathbf{X} - \mathbf{\hat{X}}||^{2}_{2} =  ||\mathbf{X} - f_{m}(\mathbf{X})||^{2}_{2}
\end{align}
%

Second, the basic model should increase the diversity of the current ensemble. Thus, the objective function $\mathcal{K}_{f_{m}}$ utilizing Equation~\ref{eq:diversity}, measures the diversity between the basic model $f_{m}(\mathbf{X})$ and the current ensemble $F(\mathbf{X})$. 
\begin{align}
    \mathcal{K}_{f_{m}} = ||f_{m}(\mathbf{X}) - F(\mathbf{X})||^{2}_{2}
\end{align}

Finally, the objective function $\mathcal{O}_{f_{m}}$ for training the basic model $f_{m}(\cdot)$ is defined as a combination of $\mathcal{J}_{f_{m}}$ and $\mathcal{K}_{f_{m}}$.
\begin{align}
    \arg\min_{\theta_{f_{m}}} \mathcal{L}_{f_{m}} = \arg\min_{\theta_{f_{m}}} \mathcal{J}_{f_{m}} - \lambda\mathcal{K}_{f_{m}},
    \label{eqn:objective}
\end{align}
where $\lambda$ is a factor that controls the importance of the diversity, and $\theta_{f_{m}}$ represents the learnable parameters of model $f_{m}(\cdot)$. 


\subsubsection{Ensemble Outlier Score}

We aggregate the outlier scores from all basic models to obtain the final outlier score. 
%
%
%
Given $M$ basic models that reconstruct the embedded time series $\mathbf{X} = \langle \mathbf{x}_{1}, \mathbf{x}_{2}, \dots, \mathbf{x}_{w} \rangle$, 
we obtain $M$ reconstructions $\hat{\mathbf{X}}^{(m)} = \langle \hat{\mathbf{x}}^{(m)}_{1}, \hat{\mathbf{x}}^{(m)}_{2}, \dots, \hat{\mathbf{x}}^{(m)}_{w} \rangle$, where $1\leq m \leq M$. 
For each vector $\mathbf{x}_{t}$ in the embedded time series $\mathbf{X}$, we obtain $M$ reconstruction errors as follows.

\begin{align}
    \langle \|\mathbf{x}_{t}-\hat{\mathbf{x}}_{t}^{(1)}\|_2^2, \|\mathbf{x}_{t}-\hat{\mathbf{x}}_{t}^{(2)}\|_2^2, \ldots, \|\mathbf{x}_{t}-\hat{\mathbf{x}}_{t}^{(M)}\|_2^2 \rangle
\end{align}
%
We use the $\mathrm{median}$ of the $M$ errors as the final outlier score of vector $\mathbf{x}_{t}$ as follows.
\begin{equation}
    OS(\mathbf{x}_{t})=\mathrm{median}\left(\|\mathbf{x}_{t}-\hat{\mathbf{x}}_{t}^{(1)}\|_2^2, \ldots, \|\mathbf{x}_{t}-\hat{\mathbf{x}}_{t}^{(M)}\|_2^2\right)
\end{equation}
We use $\mathrm{median}(\cdot)$ instead of $\mathrm{mean}(\cdot)$ because $\mathrm{median}(\cdot)$ reduces the influence of the reconstruction errors from the basic models
that overfit to the original time series~\cite{Kieu19}. 
%



{
\subsection{Unsupervised Hyperparameter Selection}\label{parameters_choice}

Hyperparameter selection is an important step in deep learning. In our setting, important hyperparmeters include, e.g., the window size $w$, the parameter transfer percentage $\beta$, and parameter $\lambda$ that controls the importance of the diversity in the loss function (cf. Equation~\ref{eqn:objective}).

Hyperparameter selection is often conducted on a validation set that is separated from the training set. We first train a model, under a specific hyperparameter setting, using the training set, and then we use the model to obtain a quality score on the validation set. Since the training and validation sets are independent, the quality score indicates how good the model, under the specific hyperparameter setting, would perform on unseen testing data. 
Based on the quality scores for different hyperparameter settings, we then select the hyperparameter setting that has the highest quality score. 

The above strategy works well in supervised settings where ground truth labels in the validation set can be used to compute directly the quality scores, e.g., mean square errors or F1 scores.
However, in an unsupervised setting, this becomes nontrivial as no ground truth labels are available.

Most existing studies for unsupervised outlier detection only provide the hyperparameters used in experiments, with no specification about how to select them. 
We summarize two strategies that exist in the literature for unsupervised outlier detection, identifying their limitation and propose a new strategy.  

First, as there are no ground truth outlier labels available in the validation set to guide the best hyperparameter, they randomly select hyperparameters~\cite{Kieu19,Chen17}. However, if we are unlucky, we may choose inadequate hyperparameters, conducting to reduced accuracy. 

Second, using small amounts of testing data for validation such that the ground truth labels in the testing data enables the computation of the quality scores. However, this strategy is no longer fully unsupervised---although the training is still unsupervised with no ground truth outlier labels, the hyperparameter selection needs the ground truth outlier labels~\cite{Cevikalp20}. 

We propose to use the reconstruction errors on the validation set as quality scores. This is fully unsupervised, since computing the reconstruction errors does not require ground truth labels. 
We may choose the hyperparameters that give the \textit{lowest} reconstruction errors. 
However, 
%
%
we observe in our experiments that this strategy often leads to sub-optimal outlier detection accuracy. When the reconstruction errors are too small, the model may overfit to the specifics in the training time series, including outliers, thus making it difficult to distinguish outliers from regular patterns. 

Our proposal uses the hyperparameters with the \textit{median} reconstruction error among all hyperparameter settings. 
%
%
Although this strategy does not guarantee that the selected hyperparameters produce a model with the best accuracy on the testing data, it often provides a good enough accuracy and often outperforms the model using the hyperparameters with the lowest reconstruction errors, as it is shown in experiments in Section~\ref{sssec:hyper}.
%
%

We employ three important hyperparameters in the model, as shown in Figure~\ref{fig:diversityOptimization}: the window size $w$, the diversity factor $\lambda$ (cf. Equation~\ref{eqn:objective}) and the fraction of parameter transfer between ensemble members $\beta$ (cf. Figure~\ref{fig:transfer}). 
The process is detailed in Algorithm~2.

\begin{algorithm}
\small
    \SetKwFunction{Split}{$\mathit{Split}$}
    \SetKwFunction{CAE}{CAE}
    \SetKwInOut{KwIn}{Input}
    \SetKwInOut{KwOut}{Output}
    \SetKwFunction{Grid}{Randomly Select ($w$, $\beta$, $\lambda$) from ($[w_1$,$w_a$], $[\beta_1$,$\beta_b$], $[\lambda_1$,$\lambda_c$])}

    \KwIn{A time series $\mathcal{T}$}
    \KwOut{Selected hyperparameter values}
    

    $\mathit{reconst}[] \leftarrow \emptyset$;
    
    $\mathcal{T}_{\mathit{training}}, \mathcal{T}_{\mathit{validation}} \leftarrow \Split(\mathcal{T})$;
    
    

    \tcc{Finding default hyperparameter values using random search.}
    \While \Grid 
    {
        
        $\mathit{Train}$ \texttt{CAE-Ensemble}$(\mathcal{T}_{\mathit{training}},w, \beta,\lambda)$;
        
        $reconst[w, \beta,\lambda] \leftarrow$ \texttt{CAE-Ensemble}$(\mathcal{T}_{\mathit{validation}})$;
    }

        
    
    ($w_{\mathit{def}}, \beta_{\mathit{def}}, \lambda_{\mathit{def}}) \leftarrow \mathit{arg}\,\mathit{median}(\mathit{reconst})$;
    
    \tcc{Choosing the Optimal Hyperparameters.}
    $w_{\mathit{opt}} \leftarrow \mathit{arg}\,\mathit{median}_{w\in[w_1, w_a]} \mathit{ReconError}(w, \beta_{\mathit{def}}, \lambda_{\mathit{def}})$; \\
    
    $\beta_{\mathit{opt}} \leftarrow \mathit{arg}\,\mathit{median}_{\beta\in[\beta_1, \beta_b]} \mathit{ReconError}(w_{\mathit{def}}, \beta, \lambda_{\mathit{def}} )$; \\
    
    $\lambda_{\mathit{opt}} \leftarrow \mathit{arg}\,\mathit{median}_{\lambda\in[\lambda_1, \lambda_c]} \mathit{ReconError} (w_{\mathit{def}}, \beta_{\mathit{def}}, \lambda)$; \\

    \KwRet{$(w_{\mathit{opt}}, \beta_{\mathit{opt}}, \lambda_{\mathit{opt}})$ }
    \caption{Unsupervised Hyperparameter Selection
    }
\end{algorithm}

The dataset is first split into training and validation sets, both without ground-truth labels. This ensures that the hyperparameter selection remains unsupervised. 
Then we define a range for each hyperparameter, i.e., $[w_1$,$w_a]$ for window size $w$. 
%
Next, we use random search~\cite{BergstraB12} to identify a specific hyperparameter combination $(w_{\mathit{def}}$, $\beta_{\mathit{def}}$, $\lambda_{\mathit{def}})$ with the median reconstruction error among all hyperparameter combinations considered in the random search. This triple is used as the default hyperparmeters. Here, we use random search instead of grid search that requires enumeration of all possible hyperparameter combinations, which is too costly. 
Finally, we identify the optimal value for each hyperparameter. To do so, we fix the other two hyperparameters to their default values and vary the hyperparameter in its range. For example, when identifying the optimal $w$, we fix $\beta$ and $\lambda$ to their default values, and vary $w$ in range $[w_1$,$w_a]$. We then choose the $w_{\mathit{opt}}$ that gives the median reconstruction error as the optimal value for hyperparameter $w$. 
}

\section{Experiments} 
\label{experiments}

We conduct extensive experiments to support the design choices for \texttt{CAE-Ensemble}. We evaluate both accuracy the efficiency.

\subsection{Experiment Setup}\label{subs:ExpSet}
\subsubsection{Data sources} \label{subs:data} We use five different multivariate public real-world time series datasets, comprising settings from different domains such as server metrics, health, and embedded systems monitoring.
All datasets, except \textit{ECG}, include a train and test sets, considering ground-truth labels indicating outliers only for testing. We do not use the labels when training, just to calculate the accuracy. For \textit{ECG}, the same set is used for both steps, ignoring the labels during training. 
For each data set, we reserve 
30\% of the training set as the validation set to enable the selection of optimal hyperparameters using the proposed median strategy. 


 \begin{itemize}
     \item \textit{ECG}\footnote{\url{https://www.cs.ucr.edu/\~eamonn/time\_series\_data\_2018/}} is a two-dimensional time series related to electrocardiogram readings for seven patients. Each time series contains 3,700-5,400 observations. The outlier ratio is 4.88\%.
     \item  \textit{SMD}\footnote{\url{https://github.com/NetManAIOps/OmniAnomaly/}} is a public dataset for server metrics consisting of 28 time series, where each has 38 dimensions. It contains 708,405 observations for training and 708,420 for testing. The outlier ratio is 4.16\%.
     \item \textit{MSL}\footnote{\label{telemanom}\url{https://github.com/khundman/telemanom}} consists on telemetry data from the Mars Curiosity Rover. It comprises 36 time series with 55 dimensions, containing 58,317 observations for training and 73,729 for testing. The outlier ratio is 9.17\%.
     \item \textit{SMAP}\textsuperscript{\ref{telemanom}} is a dataset from a Soil Moisture satellite consisting on 69 time series with 25 different variables distributed in ten subsets. In total, they consist on 138,004 training and 435,826 testing observations. The outlier ratio is 12.27\%.
     \item \textit{WADI}\footnote{\label{itrust}\url{https://itrust.sutd.edu.sg/itrust-labs_datasets/}} consists on two time series with 127 dimensions, representing a water distribution system under normal operation (1,994,172 observations) and during intrusion attacks (345,604 testing readings). The outlier ratio is 5.76\% and it is sampled every ten timestamps, given its extensive size.

 \end{itemize}

\subsubsection{Baselines}
We compare the proposed \texttt{CAE-Ensemble} with the following unsupervised outlier detection models.
 
 \begin{itemize}
    \item Isolation Forest (\texttt{IS})~\cite{Liu08}: An ensemble of randomized clustering trees that isolates outliers in sparse clusters. We use 100 base estimators for the ensemble;
    \item Local Outlier Factor (\texttt{LOF})~\cite{Breunig00}: A density clustering based method that detects outliers according to local deviations from neighbors. The number of neighbors is 20 and we use Euclidean distance;
    \item One-Class SVM (\texttt{SVM})~\cite{Scholkopf99}: A one-class classification method that employs Support Vector Machines to learn the boundary normal data points. We use a radial basis function (RBF) kernel with $\nu = 0.5$~\cite{ScholkopfSWB00}; 
    \item Moving Average Smoothing (\texttt{MAS}): A method where the values that deviates from a moving average window are likely to be considered as outliers;
    \item Autoencoder Ensemble (\texttt{AE-Ensemble})~\cite{Chen17}: 
    An ensemble that consists of feed forward autoencoders with 20\% of the connections randomly removed (cf. Table~\ref{table:summary} in Section~\ref{definitions}); 
    %
    \item Recurrent Autoencoder (\texttt{RAE})~\cite{Malhotra16}: A recurrent autoencoder using \texttt{LSTM} units to reconstruct time series via a sequence-to-sequence architecture (cf. Table~\ref{table:summary} in Section~\ref{definitions});
    \item Convolutional Autoencoder (\texttt{CAE}): The proposed convolutional sequence-to-sequence autoencoder without using an ensemble (cf. Section~\ref{subs:ConvSeq});
    %
    %
    \item Correlation Matrices Recurrent Autoencoder (\texttt{MSCRED})~\cite{Zhang19}: A state-of-the-art method for multivariate time series outlier detection that uses an autoencoder to reconstruct correlation matrices instead of using the time series directly. Matrices have length 16 with 5 steps in-between;
    \item Variational Recurrent Autoencoder (\texttt{RNNVAE})~\cite{Soelch16}: The model establishes a stochastic latent component in the autoencoder for learning a distribution to improve the reconstruction output. The hidden and stochastic spaces are 64 with a regularization of 0.0001;
    \item Temporal Variational Autoencoder (\texttt{OMNIANOMALY})~\cite{Su19}: The method extends the previous variational modeling with an additional component to capture temporal dependencies in the context of stochastic variables. The hidden space is 32 with 16 stochastic variables and a regularization of 0.0001;
    \item  Recurrent Autoencoder Ensembles (\texttt{RAE-Ensemble})~\cite{Kieu19}: A state-of-the-art recurrent autoencoder ensemble (cf. Table~\ref{table:summary} in Section~\ref{definitions}), 
    where 20\% of the skip
    connections are randomly dropped.
    
 \end{itemize}

\subsubsection{Evaluation Metrics} 
We consider two categories of metrics. 
%
%

\noindent
\textbf{All thresholds:} We consider a setting where prior knowledge on selecting a specific threshold is not provided. In this case, 
we use two metrics that consider all possible thresholds~\cite{Chen17,Kieu19}---Area Under the Curve of Precision-Recall (\textit{PR}) 
and Area Under the Curve of Receiver Operating Characteristic (\textit{ROC})~\cite{Sammut17}. 
Both metrics are defined in terms of true positives (\textit{TP}) (i.e., outliers predicted as outliers), true negatives (\textit{TN}) (i.e., inliers predicted as inliers), false positives (\textit{FP}) (i.e., inliers predicted as outliers), and false negatives (\textit{FN}) (i.e., outliers predicted as inliers). The difference between \textit{PR} and \textit{ROC} is that \textit{PR} disregards \textit{TN}.
The higher the \textit{PR} and \textit{ROC} are, the more accurate a method is.

\noindent
\textbf{Specific thresholds:} We consider a setting where a specific threshold is chosen and 
we measure \textit{Precision}, \textit{Recall}, and \textit{F1}~\cite{Sammut17} with the ground-truth outlier labels using the specific threshold. Higher values are desired. 
Here, we consider two settings. First, we select a threshold, among all possible thresholds, that gives the ``best'' \textit{F1}, i.e., the best possible threshold~\cite{Xu18,Su19}. 
%
We use this threshold to report the best \textit{F1}, along with the corresponding \textit{Precision} and \textit{Recall}. 
%
%
Second, we consider a specific threshold that is decided by the prior knowledge on the outlier ratio. In particular, if we know that a data set has K\% of outliers, we choose a threshold such that there are K\% of data whose ouliter scores are higher than the threshold.


\subsubsection{Hyperparameters} \label{sssec:hyperWindows}

\begin{figure}[ht!]
    \centering
    \includegraphics[width=0.9\linewidth]{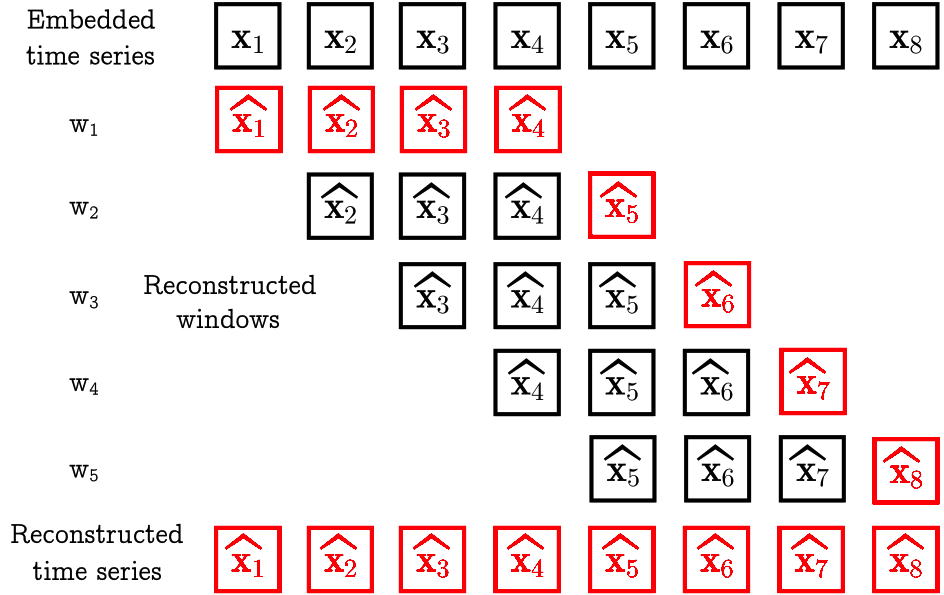}
    \caption{Outlier scores from different windows. 
    }
    \label{fig:outlier_scores}
    \vspace{-1em}
\end{figure}

The transfer $\beta$ and diversity $\lambda$ parameters (see Equation~\ref{eqn:objective}) are set according to the median strategy presented in Section~\ref{parameters_choice}, evaluating the cases $\beta = i/10, i \in [1,9]$ and $\lambda = 2^j, j \in [0,6]$ for each data set. Then, each time series is divided into windows of size $w$ with $w-1$ overlapped observations with respect to the previous window. 
For each window, we consider the reconstruction of the last observation in the window when calculating the outlier scores, except in the case of the first window, where all reconstructions are considered. 
This is shown the red boxes in Figure~\ref{fig:outlier_scores}. The first window consider all reconstructed observations to compute the outlier scores. Then, for each of the remaining windows, we use only the reconstructed observation from the last timestamp to compute the outlier score.  

The $w$ is also chosen according to the median strategy, considering $w = 2^k, k \in [2,8]$. 
%
%
The selected hyperparameters for different data sets are shown in Table~\ref{table:parameters}. The details of hyperparameter selection are offered in Sections~\ref{sssec:hyper} and~\ref{sssec:hyper2}. 

\begin{table}[ht!]
    \centering
    \small
    \caption{Selected hyperparameters, median strategy.}
    \label{table:parameters}
    \begin{tabular}{ |c|c|c|c|c|c| } 
        \hline
         &  \textit{ECG} & \textit{MSL} & \textit{SMAP} & \textit{SMD} & \textit{WADI}\\ 
        \hline
$\beta$ & 0.5 & 0.7 & 0.9 & 0.2 & 0.5 \\
$\lambda$ & 2 & 16 & 2 & 32 & 1 \\
$w$ & 16 & 16 & 16 & 32 & 32 \\
        \hline
    \end{tabular}
\end{table}


 \subsubsection{Implementation Details} 
We set the number of convolution layers in \texttt{CAE} to 10 for both the encoder and the decoder, with the kernel size for the convolution operator in all layers set to 3. The dimension of embedded vector $D'$ is set to 256, and the batch sizes to 64. 
By default we use 8 basic models, with a new basic model for every 50 epochs. This setting ensures that a newly generated basic model is learned from previously well-trained basic models. We also vary the number of basic models in Section~\ref{sssec:accuracy}. 
We use Adam~\cite{Kingma15}, a Stochastic Gradient Descent variant, as the optimizer. The learning rate is set to 0.001. 


The experiments are run on a Ubuntu 18.04 server with a 10-cores CPU Intel Xeon W-2155, 128GB RAM, and two GPU NVIDIA TITAN RTXs. The code was written in Python 3.9.1 using PyTorch 1.7.1. 
The source code is publicly available at 
{\url{https://github.com/d-gcc/CAE-Ensemble}}.

\subsection{Experiment Results}
\subsubsection{Accuracy} We report the accuracy results for the five datasets, respectively, as well as the average accuracy over all datasets. We report all-threshold metrics \textit{PR} and \textit{ROC} and threshold-dependent metrics \textit{Precision}, \textit{Recall}, and \textit{F1}, when using the threshold that gives the best \textit{F1}. We use bold and underline to highlight the highest and second highest values. 


\renewcommand{\tabcolsep}{2.5pt}

\begin{table*}[ht!]
    \small
    \centering
    \caption{\textit{ECG}, \textit{SMD}, and \textit{MSL} accuracy results.}
    \label{table:ECG_SMD_MSL}
    \begin{tabular}{ |l|c|c|c|c|c|c|c|c|c|c|c|c|c|c|c| } 
    \hline
    \textit{Data set} & \multicolumn{5}{c|}{\textit{ECG}} & \multicolumn{5}{c|}{\textit{SMD}} & \multicolumn{5}{c|}{\textit{MSL}}\\
        \hline
        \textbf{Model} &  \textit{Precision} &  \textit{Recall} &  \textit{F1} & \textit{PR}  & \textit{ROC} &
        \textit{Precision} &  \textit{Recall} &  \textit{F1} & \textit{PR}  & \textit{ROC} &
        \textit{Precision} &  \textit{Recall} &  \textit{F1} & \textit{PR}  & \textit{ROC} \\
        \hline
\texttt{ISF} & 0.0543 & \textbf{0.7199} & 0.0999 & 0.0501 & 0.5062 
    & 0.0880 & \underline{0.4571} & 0.1079 & 0.0591 & 0.5066
    & 0.1553 & 0.6512 & 0.1895 & 0.1085 & 0.5036 \\
    
\texttt{LOF} & 0.0539 & 0.6539 & 0.0962 & 0.0500 & 0.4912
    & 0.2494 & 0.2571 & 0.1764 & 0.1203 & 0.5695
    & 0.2463 & 0.5316 & 0.2358 & 0.1431 & 0.5268 \\
\multicolumn{1}{|l|}{\texttt{MAS}} & 0.0670 & 0.6276 & 0.1159 & 0.0578 & 0.5342
    & \underline{0.4720} & 0.4099 & 0.3716 & \underline{0.3253} & \textbf{0.7520}
    & \textbf{0.2959} & 0.5537 & 0.2525 & 0.1595 & 0.5469 \\
\multicolumn{1}{|l|}{\texttt{OCSVM}} & 0.0825 & 0.4987 & 0.1309 & 0.0588 & 0.5342
    & 0.3414 & 0.2944 & 0.2626 & 0.1927 & 0.5783 
    & \underline{0.2847} & 0.5149 & 0.2616 & 0.1581 & 0.5629 \\
         \hline
\multicolumn{1}{|l|}{\texttt{MSCRED}} & 0.1789 & \underline{0.6651} & \underline{0.2303} & 0.1055 & 0.5166
    & 0.0631 & \textbf{0.7719} & 0.1100 & 0.0395 & 0.5000
    & 0.1243 & \textbf{0.7747} & 0.1874 & 0.1166 & 0.5072 \\

\multicolumn{1}{|l|}{\texttt{OMNIANOMALY}} & \underline{0.2220} & 0.4938 & 0.2042 & \underline{0.1409} & 0.5584
    & 0.2432 & 0.3328 & 0.2110 & 0.1503 & 0.6148 
    & 0.1936 & 0.6297 & 0.2414 & 0.1609 & 0.5429 \\

\multicolumn{1}{|l|}{\texttt{RNNVAE}} & 0.1768 & 0.4222 & 0.1439 & 0.0895 & 0.5500
    & 0.4334 & 0.3194 & 0.3045 & 0.2406 & 0.6917
    & 0.1641 & 0.5639 & 0.2125 & 0.1378 & 0.5335 \\

\multicolumn{1}{|l|}{\texttt{AE-Ensemble}} & 0.1583 & 0.5398 & 0.1907 & 0.1302 & \textbf{0.5952}
    & 0.3713 & 0.3709 & 0.2832 & 0.2349 & 0.6823 
    & 0.1775 & \underline{0.6936} & 0.2424 & 0.1404 & 0.5360 \\

\multicolumn{1}{|l|}{\texttt{RAE}} & 0.1297 & 0.5394 & 0.1669 & 0.0936 & \underline{0.5922}
    & 0.4466 & 0.3037 & 0.3078 & 0.2424 & 0.6836
    & 0.2069 & 0.6091 & 0.2423 & 0.1503 & 0.5575 \\

\multicolumn{1}{|l|}{\texttt{RAE-Ensemble}} & 0.2003 & 0.5838 & 0.1864 & 0.1176 & 0.5372
    & 0.4684 & 0.3318 & 0.3332 & 0.2639 & 0.6998
    & 0.2085 & 0.5633 & 0.2495 & 0.1572 & 0.5714 \\

         \hline
\multicolumn{1}{|l|}{\texttt{CAE}} & 0.1919 & 0.4574 & 0.1954 & 0.1297 & 0.5633
    & 0.4625 & 0.3804 & \textbf{0.3895} & \textbf{0.3299} & \underline{0.7416}
    & 0.2223 & 0.5273 & \underline{0.2649} & \textbf{0.1641} & \underline{0.5843} \\

\multicolumn{1}{|l|}{\texttt{CAE-Ensemble}} & \textbf{0.2522} & 0.4924 & \textbf{0.2521} & \textbf{0.1887} & 0.5715
    & \textbf{0.4924} & 0.3739 & \underline{0.3770} & 0.3246 & 0.7375 
    & 0.2501 & 0.5343 & \textbf{0.2713} & \underline{0.1633} & \textbf{0.5963} \\

        \hline

    \end{tabular}
\end{table*}
\begin{table*}[ht!]
    \small
    \centering
    \caption{\textit{SMAP}, \textit{WADI}, and \textit{Overall} accuracy results.}
    \label{table:SMAP_WADI_Overall}
    \begin{tabular}{ |l|c|c|c|c|c|c|c|c|c|c|c|c|c|c|c| } 
    \hline
    \textit{Data set} & \multicolumn{5}{c|}{\textit{SMAP}} & \multicolumn{5}{c|}{\textit{WADI}} & \multicolumn{5}{c|}{\textit{Overall}}\\
        \hline
        \multicolumn{1}{|l|}{\textbf{Model}} &  \textit{Precision} &  \textit{Recall} &  \textit{F1} & \textit{PR}  & \textit{ROC} &
        \textit{Precision} &  \textit{Recall} &  \textit{F1} & \textit{PR}  & \textit{ROC} &
        \textit{Precision} &  \textit{Recall} &  \textit{F1} & \textit{PR}  & \textit{ROC} \\
        \hline
\multicolumn{1}{|l|}{\texttt{ISF}} 
& 0.1396 & 0.5298 & 0.1986 & 0.1300 & 0.4979
& 0.0667 & \underline{0.4765} & 0.1170 & 0.0610 & 0.5248
& 0.1008 & \underline{0.5669} & 0.1426 & 0.0818 & 0.5078 \\
    
\multicolumn{1}{|l|}{\texttt{LOF}} 
& 0.2261 & 0.5178 & 0.2027 & 0.1289 & 0.5005
& 0.0736 & 0.3155 & 0.1193 & 0.0702 & 0.5284
 & 0.1698 & 0.4552 & 0.1661 & 0.1025 & 0.5233 \\
    
\multicolumn{1}{|l|}{\texttt{MAS}} 
& 0.2819 & 0.5174 & 0.2542 & 0.1655 & 0.5233
& 0.2586 & 0.1555 & 0.1942 & 0.1490 & 0.5788 
 & 0.2751 & 0.4528 & 0.2377 & 0.1714 & 0.5870 \\
    
\multicolumn{1}{|l|}{\texttt{OCSVM}} 
& 0.2561 & 0.5722 & 0.2302 & 0.1461 & 0.4924
& 0.0980 & 0.2955 & 0.1472 & 0.1192 & 0.5754 
 & 0.2125 & 0.4351 & 0.2065 & 0.1350 & 0.5487 \\

         \hline
\multicolumn{1}{|l|}{\texttt{MSCRED}} 
& 0.1266 & \textbf{0.8199} & 0.1914 & 0.1028 & 0.4403
& 0.1382 & \textbf{0.8590} & 0.2377 & 0.0993 & \underline{0.6730}
& 0.1262 & \textbf{0.7781} & 0.1913 & 0.0927 & 0.5274 \\

\multicolumn{1}{|l|}{\texttt{OMNIANOMALY}} 
& 0.2307 & 0.6222 & 0.2681 & 0.1556 & 0.5402
& 0.2996 & 0.3976 & \textbf{0.3404} & 0.1723 & \textbf{0.7261}
& 0.2378 & 0.4952 & 0.2530 & 0.1560 & 0.5965 \\

\multicolumn{1}{|l|}{\texttt{RNNVAE}} 
& 0.1622 & 0.5646 & 0.1971 & 0.1192 & 0.5119
& 0.2881 & 0.3147 & \underline{0.2867} & \underline{0.1734} & 0.5739
& 0.2449 & 0.4370 & 0.2289 & 0.1521 & 0.5722 \\

\multicolumn{1}{|l|}{\texttt{AE-Ensemble}} 
& 0.3134 & 0.5895 & 0.2939 & 0.1780 & 0.5496
& 0.1619 & 0.2398 & 0.1928 & 0.0911 & 0.5102
& 0.2404 & 0.4727 & 0.2379 & 0.1498 & 0.6078 \\

\multicolumn{1}{|l|}{\texttt{RAE}} 
& 0.2071 & 0.6316 & 0.2381 & 0.1476 & 0.5390
& 0.2118 & 0.2799 & 0.2342 & 0.1150 & 0.6667
& 0.2365 & 0.4867 & 0.2406 & 0.1549 & 0.5747 \\

\multicolumn{1}{|l|}{\texttt{RAE-Ensemble}} 
& 0.2603 & \underline{0.6604} & 0.2529 & 0.1628 & 0.5716
& \underline{0.2999} & 0.2535 & 0.2707 & 0.1580 & 0.6516
& \underline{0.2875} & 0.4786 & 0.2585 & 0.1719 & 0.6063 \\

         \hline
\multicolumn{1}{|l|}{\texttt{CAE}} 
& \underline{0.3175} & 0.5912 & \underline{0.3170} & \underline{0.2135} & \underline{0.5892}
& 0.2350 & 0.3019 & 0.2004 & 0.1243 & 0.5994
& 0.2858 & 0.4516 & \underline{0.2735} & \underline{0.1923} & \underline{0.6156} \\

\multicolumn{1}{|l|}{\texttt{CAE-Ensemble}} 
& \textbf{0.3387} & 0.6187 & \textbf{0.3327} & \textbf{0.2223} & \textbf{0.6080}
& \textbf{0.5006} & 0.1995 & 0.2853 & \textbf{0.1911} & 0.6023
& \textbf{0.3668} & 0.4438 & \textbf{0.3037} & \textbf{0.2180} & \textbf{0.6231} \\
\hline

        \hline
    \end{tabular}
\end{table*}

The results of \textit{ECG} are shown in the first section of Table~\ref{table:ECG_SMD_MSL}. 
The neural network based methods outperform the non-neural network based methods in almost all metrics. The relatively high \textit{Recall} for \texttt{ISF} comes with a price with low \textit{Precision}, which means that many instances that are not outliers are mistakenly considered as outliers. 
On average, \textit{CAE-Ensemble} achieves the best result for \textit{Precision}, \textit{F1}, and \textit{PR} with a competitive \textit{ROC} case, showing a very good performance overall in comparison to the baselines.  

The results for \textit{SMD} are shown 
 in the middle of Table~\ref{table:ECG_SMD_MSL}.
where \textit{CAE} and \textit{CAE-Ensemble} achieve higher accuracy on most of the metrics, in particular, \textit{Precision}, \textit{F1}, and \textit{PR}, demonstrating the model functionality in server-related data sets. The high \textit{Recall} in \textit{MSCRED} reflects the case detailed before, i.e., with the price of low \textit{Precision}. 

For \textit{MSL}, in the last part of Table~\ref{table:ECG_SMD_MSL}, 
and \textit{SMAP} in the first part of Table~\ref{table:SMAP_WADI_Overall}, 
our proposals show a great performance on the all-threshold metrics \textit{PR} and \textit{ROC} and also on the threshold-dependent metric \textit{F1} that represents a balance between \textit{Precision} and \textit{Recall}, compared to all baselines. Then, 
for \textit{WADI}, in the middle section of Table~\ref{table:SMAP_WADI_Overall}, 
the proposed models also present a robust performance, suggesting that the proposals also work in an application domain of water distribution.


The last part of Table~\ref{table:SMAP_WADI_Overall} 
summarizes the overall performance for all the time series. 
The results show that our method outperforms all the other baselines w.r.t. \textit{Precision}, \textit{F1}, \textit{PR}, and \textit{ROC}, suggesting that \texttt{CAE-Ensemble} is a very competitive solution in terms of quality. For the \textit{Recall} metric, the better performance shown by \texttt{MSCRED} and \texttt{ISF} come with very low \textit{Precision}, making them not competitive in real application scenarios. This occurs when such methods are overly pessimistic and identify many observations as outliers, although some observations are not outliers. This produces more true positives at the cost of obtanining many false positives. This gives 
high \textit{Recall} but low \textit{Precision}. In addition, \texttt{CAE-Ensemble} performs not so well on \textit{WADI}, where it has a very low \textit{Recall} and thus low \textit{F1} and \textit{ROC}. This is due to how the ground-truth is provided in the data set---the ground-truth marks all observations in long  intervals as outliers, whereas only a few observations in the middle of the long intervals are real outliers. Although \texttt{CAE-Ensemble} is often able to detect the real outliers, it still has a low recall. 

\begin{figure}[ht!]
    \centering
    \includegraphics[width=0.8\linewidth]{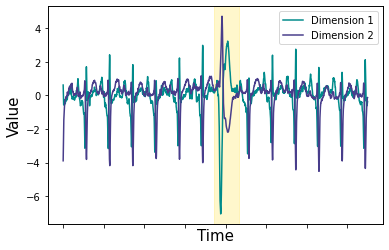}
    \caption{An example of a two dimensional time series from the ECG data set (subset chf13). The yellow region indicates a ground-truth outlier interval.
    }
    \label{fig:series_ecg}
    \vspace{-1em}
\end{figure}

We have conducted a deep analysis on the \textit{Recall} scores. The main reason lies in that the ground-truth labels often mark outlier intervals, not outlier observations, although an outlier interval is often caused by only a few outlier observations in the interval. 
Figure~\ref{fig:series_ecg} shows an example where the yellow region indicates an outlier interval and where all observations in the interval are marked as ground-truth outliers. However, some observations in the interval are clearly more different than others in the interval, e.g., the two peaks.

\begin{figure}[ht!]
    \centering
    \includegraphics[width=0.8\linewidth]{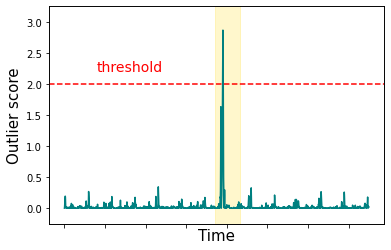}
    \caption{Outlier scores, \texttt{CAE-Ensemble}.
    }
    \label{fig:score_cae}
    \vspace{-1em}
\end{figure}

We then visualize the outlier scores of  our proposal  \texttt{CAE-Ensemble} in Figure~\ref{fig:score_cae}.
For \texttt{CAE-Ensemble}, in the yellow region, only a few points have very high outlier scores, i.e., larger than the threshold. These are aligned well with the two peaks and are thus considered as outliers. Thus, the \textit{Recall} is low and the \textit{Precision} is high. 


Due to the space limitation, in the following experiments, we only report results on two data sets, \textit{ECG} and \textit{SMAP}, as other data sets show similar results. 



\subsubsection{Outlier Score Threshold Sensitivity.} 
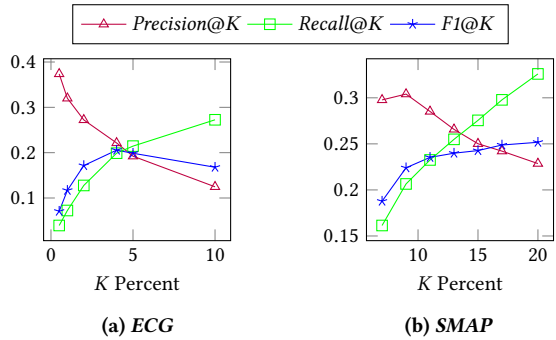
\begin{figure}[ht!]
\small
\centering 
  \begin{subfigure}[b]{0.48\linewidth}
  \hspace{0.2cm}
  \begin{tikzpicture}
    \begin{axis}[
        xlabel=\textit{K} Percent,
        width=\linewidth,
        height=0.55*\axisdefaultheight,
        legend style={at={(1.32,1.30)},anchor=north,legend columns=-1}]
        \addplot[purple, mark=triangle] table[x=t, y=a] {figures/data/K_ECG.txt};
        \addlegendentry{\textit{Precision}@$K$}
        \addplot[green, mark=square] table[x=t, y=b] {figures/data/K_ECG.txt};
        \addlegendentry{\textit{Recall}@$K$}
        \addplot[blue, mark=star] table[x=t, y=c] {figures/data/K_ECG.txt};
        \addlegendentry{\textit{F1}@$K$}
        \end{axis}
    \end{tikzpicture}
    \caption{\textit{ECG}} \label{subfig:KECG}  
  \end{subfigure}
\begin{subfigure}[b]{0.48\linewidth}
  \hspace{0.2cm}
  \begin{tikzpicture}
    \begin{axis}[
        xlabel=\textit{K} Percent,
        width=\linewidth,
        yticklabel style={
        /pgf/number format/fixed,
        /pgf/number format/precision=2},
        scaled y ticks=false,
        height=0.55*\axisdefaultheight,
        legend style={at={(1.32,1.30)},anchor=north,legend columns=-1}]
        \addplot[purple, mark=triangle] table[x=t, y=a] {figures/data/K_SMAP_P.txt};
        \addplot[green, mark=square] table[x=t, y=b] {figures/data/K_SMAP_P.txt};
        \addplot[blue, mark=star] table[x=t, y=c] {figures/data/K_SMAP_P.txt};
        \end{axis}
    \end{tikzpicture}
    \caption{\textit{SMAP}} \label{subfig:KSMAP}  
\end{subfigure}
\caption{Outlier score threshold sensitivity with top $K$\% largest outlier scores.}
\label{fig:topK}
\end{figure}

To characterize the metrics related to specific thresholds, in Figure~\ref{fig:topK}, we show their evolution as we select the top $K$ percentage of the largest outlier scores as the threshold and consider the observations which have larger outlier scores as outliers. 

For \textit{ECG}, Figure~\ref{subfig:KECG} shows convergence at $K=5$ percent, which is close to the actual outlier ratio (i.e., 4.88\% cf. Section~\ref{subs:data}).  
%
\textit{SMAP} includes multiple subsets whose outlier ratios differ significantly, ranging from 0.8\% to 21.9\%.  Figure~\ref{subfig:KSMAP} shows a representative subset from \textit{SMAP} with an outlier ratio of 12\%, which is close to the average outlier ratio of \textit{SMAP} (cf. Section~\ref{subs:data}). 

Both figures suggest that when the outlier ratio is available, choosing it as the $K$ percentage is a good choice. 

\subsubsection{Ablation Study} 
To evaluate our design choices, we study the effect of each module of \texttt{CAE-Ensemble} by removing model components. Specifically, we (1) remove the attention module (\textit{No attention}); (2) remove the parameter transfer learning and without using the diversity metric on the objective function in the ensemble such that all basic models are trained independently (\textit{No diversity}); (3) disregard the ensemble, i.e., using a single basic model \texttt{CAE} (\textit{No ensemble}), and (4) remove the re-scaling in the pre-processing step and use the raw time series directly (\textit{No re-scaling}).

\begin{table}[ht!]
    \small
    \centering
    \caption{Ablation study, \textit{ECG} and \textit{SMAP}.}
    \label{table:BothAblation}
    \begin{tabular}{ |c|l|c|c|c|c|c| } 
        \hline
        \textbf{} & & \textit{Precision} &  \textit{Recall} &  \textit{F1} & \textit{PR}  & \textit{ROC} \\ 
        \hline
\parbox[t]{2mm}{\multirow{5}{*}{\rotatebox[origin=c]{90}{\textit{ECG}}}} &\texttt{No attention} & 0.1440 & 0.4809 & 0.1840 & 0.1037 & 0.5606 \\
& \texttt{No diversity} & 0.1683 & 0.4714 & 0.1819 & 0.1244 & \textbf{0.5939} \\
& \texttt{No ensemble} & 0.1919 & 0.4574 & 0.1954 & 0.1297 & 0.5633 \\
& \texttt{No re-scaling} & 0.1806 & 0.4819 & 0.1741 & 0.1130 & 0.5379 \\
& \texttt{\texttt{CAE-Ensemble}} & \textbf{0.2522} & \textbf{0.4924} & \textbf{0.2521} & \textbf{0.1887} & 0.5715 \\
        \hline

\parbox[t]{2mm}{\multirow{5}{*}{\rotatebox[origin=c]{90}{\textit{SMAP}}}} & \texttt{No attention} & 0.3290 & 0.5763 & 0.3049 & 0.1957 & 0.5605 \\
& \texttt{No diversity} & 0.3241 & 0.5841 & 0.3210 & 0.2186 & 0.5832 \\
& \texttt{No ensemble} & 0.3175 & 0.5912 & 0.3170 & 0.2135 & 0.5892 \\
& \texttt{No re-scaling} & 0.3252 & 0.5689 & 0.2872 & 0.1938 & 0.5666 \\
& \texttt{\texttt{CAE-Ensemble}} & \textbf{0.3387} & \textbf{0.6187} & \textbf{0.3327} & \textbf{0.2223} & \textbf{0.6080} \\
        \hline
    \end{tabular}
\end{table}

Table~\ref{table:BothAblation} shows that the model with all components achieves the best results in for every modification on almost all metrics. The only exception is the \textit{ROC} metric for \textit{ECG} when no diversity is considered. The diversity may introduce a few noisy false positives at some thresholds, which adversely affects the average performance since that data set is small. 

Next, we evaluate the diversity metric according to Equation~\ref{eq:divensem} for the two ensemble models. 
%
%
Table~\ref{table:diverse_ensemble} shows the results, where a higher value indicates an ensemble has more diverse based models, which is more desirable~\cite{Aggarwal17}. 
As expected, the full \texttt{CAE-Ensemble} model that considers diversity in the loss function achieves higher diversity. In contrast, for \texttt{No Diversity} ensemble, the basic models are trained independently using different random seeds to initialize their model parameters, which still results in different basic models, but less diverse ones.

\begin{table}[ht!]
    \centering
    \small
    \caption{Quantifying the diversity.}
    \label{table:diverse_ensemble}
    \begin{tabular}{ |l|c|c|c|c| } 
        \hline
         &  \textit{ECG} & \textit{SMAP}\\ 
        \hline
        \texttt{No Diversity} & 57.0118 & 16.3409 \\
        \texttt{CAE-Ensemble} & \textbf{94.7425} & \textbf{52.0796} \\
        \hline
    \end{tabular}
\end{table}

\subsubsection{Hyperparameters Selection for $\beta$ and $\lambda$}\label{sssec:hyper}
\begin{figure}[ht!] 
\small
\centering 
    \begin{subfigure}[t]{0.48\linewidth}
    \hspace{0.03\linewidth}
        \edef\windows{"0.7","0.3","0.8","0.1","0.5","0.4",
        "0.9","0.6","0.2"}
        \begin{tikzpicture}
        \begin{axis}[
            width=\linewidth,
            height=0.55*\axisdefaultheight,
            ymax=0.70,
            ymin=0.05,
            xtick={2,5,8},
            xticklabels={0.03,0.08,0.11},            
            name=bottom axis,
            xlabel=Reconstruction errors,
            legend style={at={(1.3,1.3)},anchor=north,legend columns=-1}]
        ]
            \addplot[blue] table[x=t, y=b] {figures/data/Beta_ECG.txt};
            \addlegendentry{\textit{ROC}}
            \addplot[red, nodes near coords=\pgfmathsetmacro{\winstring}{{\windows}[\coordindex]}\winstring, nodes near coords style={text=black,font=\scriptsize}] table[x=t, y=a] {figures/data/Beta_ECG.txt};
            \addlegendentry{\textit{PR}}
            \draw[line width=0.4mm](1.5,0.177) -- (2.5,0.177);
            \addplot[gray, dashed] coordinates {(5,0.05)(5,0.70)};
        \end{axis}

    \end{tikzpicture}
    \caption{$\beta$, \textit{ECG}} \label{subfig:betaECG}  
    \end{subfigure}
    \begin{subfigure}[t]{0.48\linewidth}
    \hspace{0.03\linewidth}
        \edef\windows{"4","1","64","2","32","8","16"}
        \begin{tikzpicture}
        \begin{axis}[
            width=\linewidth,
            ymax=0.70,
            ymin=0.05,
            xtick={2,4,6},
            xticklabels={0.07,0.10,0.14},   
            height=0.55*\axisdefaultheight,
            name=bottom axis,
            xlabel=Reconstruction errors,
        ]
            \addplot[red, nodes near coords=\pgfmathsetmacro{\winstring}{{\windows}[\coordindex]}\winstring, nodes near coords style={text=black,font=\footnotesize}] table[x=t, y=a] {figures/data/Lambda_ECG.txt};
            \addplot[blue] table[x=t, y=b] {figures/data/Lambda_ECG.txt};
            \addplot[gray, dashed] coordinates {(4,0.05)(4,0.70)};
        \end{axis}
    \end{tikzpicture}
    \caption{$\lambda$, \textit{ECG}} \label{subfig:lambdaECG}  
    \end{subfigure}
    
    \begin{subfigure}[b]{0.48\linewidth}
    \hspace{0.03\linewidth}
        \edef\windows{"0.4","0.2","0.7","0.3","0.9","0.6",
        "0.1","0.8","0.5"}
        \begin{tikzpicture}
        \begin{axis}[
            width=\linewidth,
            height=0.55*\axisdefaultheight,
            ymax=0.70,
            ymin=0.10,
            xtick={2,5,8},
            xticklabels={469,507,709},            
            name=bottom axis,
            xlabel=Reconstruction errors,
        ]
            \addplot[red, nodes near coords=\pgfmathsetmacro{\winstring}{{\windows}[\coordindex]}\winstring, nodes near coords style={text=black,font=\tiny}] table[x=t, y=a] {figures/data/Beta_SMAP.txt};
            \addplot[blue] table[x=t, y=b] {figures/data/Beta_SMAP.txt};
            \addplot[gray, dashed] coordinates {(5,0.10)(5,0.70)};
        \end{axis}

    \end{tikzpicture}
    \caption{$\beta$, \textit{SMAP}} \label{subfig:betaSMAP}  
    \end{subfigure}
    \begin{subfigure}[b]{0.48\linewidth}
    \hspace{0.03\linewidth}
        \edef\windows{"8","4","16","2","32","64","1"}
        \begin{tikzpicture}
        \begin{axis}[
            width=\linewidth,
            ymax=0.70,
            ymin=0.10,
            xtick={2,4,6},
            xticklabels={439,492,514},   
            height=0.55*\axisdefaultheight,
            name=bottom axis,
            xlabel=Reconstruction errors,
        ]
            \addplot[red, nodes near coords=\pgfmathsetmacro{\winstring}{{\windows}[\coordindex]}\winstring, nodes near coords style={text=black,font=\footnotesize}] table[x=t, y=a] {figures/data/Lambda_SMAP.txt};   \addplot[blue] table[x=t, y=b] {figures/data/Lambda_SMAP.txt};
            \addplot[gray, dashed] coordinates {(4,0.10)(4,0.70)};
        \end{axis}

    \end{tikzpicture}
    \caption{$\lambda$, \textit{SMAP}} \label{subfig:lambdaSMAP}  
    \end{subfigure}
\caption{{Hyperparameters selection---the numbers over the \textit{PR} curve are candidate values for $\beta$ and $\lambda$. 
}}
\label{fig:hyperparametersECG}
\end{figure}
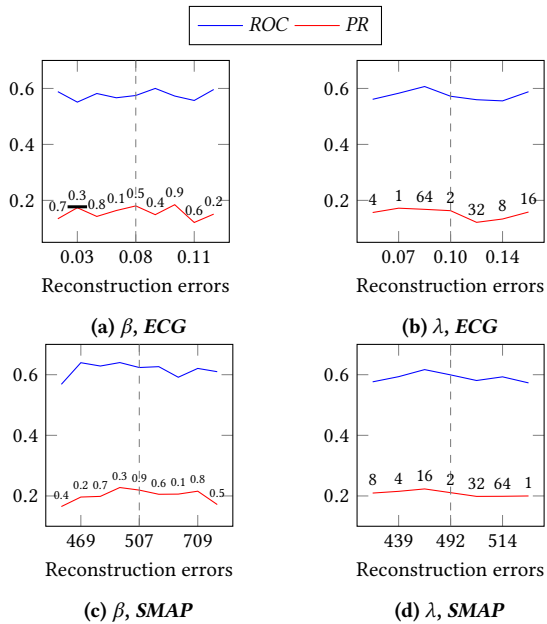

Figure~\ref{fig:hyperparametersECG} shows the $\beta$ and $\lambda$ parameter choices for the \textit{ECG} and \textit{SMAP} datasets according to the median reconstruction error strategy. Here, we consider metrics  
\textit{PR} and \textit{ROC} to evaluate the accuracy of the outlier detection. Note that computing \textit{PR} and \textit{ROC} requires ground-truth outlier labels, which are unavailable during validation. 

The results in Figure~\ref{fig:hyperparametersECG} are ordered by the associated reconstruction errors. Then, the median reconstruction error is marked by the dashed vertical line in the middle of each figure. The numbers along the \textit{PR} curve show specific values for the considered hyperparameter. 
For example, in Figure~\ref{subfig:betaECG}, when $\beta=0.3$ (underlined), its corresponding reconstruction error is 0.03, the corresponding \textit{PR} is a bit more than 0.17, and the corresponding \textit{ROC} is below 0.57. 

Figure~\ref{fig:hyperparametersECG} shows that the selected $\beta$ and $\lambda$ do not represent the optimal values that give the highest \textit{PR} or \textit{ROC}. 
However, the median strategy achieves good enough \textit{PR} and \textit{ROC}. Compared to the strategy which always selects the values with the smallest reconstruction errors, the median strategy often achieves higher \textit{PR} and \textit{ROC}. 

The results for $\beta$ in Figures~\ref{subfig:betaECG} and \ref{subfig:betaSMAP} are relatively unstable, showing sudden changes mainly for \textit{ECG}. However, the median case is balanced between the best and worst cases, suggesting that it is a robust strategy. 

The $\lambda$ parameter, in Figures~\ref{subfig:lambdaECG} and \ref{subfig:lambdaSMAP}, seems to have a trend, where the better results are achieved just before the median validation error. That observation is useful for defining enhanced criteria over using the validation error for selecting that specific parameter. Again, the median strategy is a robust and good enough strategy. 



\subsubsection{Hyperparameters Selection for Window Size $w$} 
\label{sssec:hyper2}
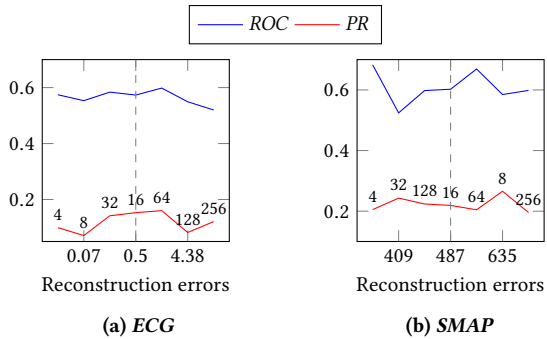
\begin{figure}[ht!] 
\small
\centering 
    \begin{subfigure}[b]{0.48\linewidth}
    \hspace{0.03\linewidth}
        \edef\windows{"4","8","32","16","64","128","256"}
        \begin{tikzpicture}
        \begin{axis}[
            width=\linewidth,
            height=0.55*\axisdefaultheight,
            ymax=0.70,
            ymin=0.05,
            xtick={2,4,6},
            xticklabels={0.07,0.5,4.38},
            name=bottom axis,
            xlabel=Reconstruction errors,
            legend style={at={(1.3,1.3)},anchor=north,legend columns=-1}]
        ]
            \addplot[blue] table[x=t, y=b] {figures/data/Window_ECG.txt};
            \addlegendentry{\textit{ROC}}
            \addplot[red, nodes near coords=\pgfmathsetmacro{\winstring}{{\windows}[\coordindex]}\winstring, nodes near coords style={text=black,font=\footnotesize}] table[x=t, y=a] {figures/data/Window_ECG.txt};
            \addlegendentry{\textit{PR}}
            \addplot[gray, dashed] coordinates {(4,0.05)(4,0.70)};
        \end{axis}

    \end{tikzpicture}
    \caption{\textit{ECG}} \label{subfig:WindowECG}  
    \end{subfigure} 
    \begin{subfigure}[b]{0.48\linewidth}
    \hspace{0.03\linewidth}
        \edef\windows{"4","32","128","16","64","8","256"}
        \begin{tikzpicture}
        \begin{axis}[
            width=\linewidth,
            ymax=0.70,
            ymin=0.10,
            xtick={2,4,6},
            xticklabels={409,487,635},
            height=0.55*\axisdefaultheight,
            name=bottom axis,
            xlabel=Reconstruction errors,
        ]
            \addplot[red, nodes near coords=\pgfmathsetmacro{\winstring}{{\windows}[\coordindex]}\winstring, nodes near coords style={text=black,font=\footnotesize}]  table[x=t, y=a] {figures/data/Window_SMAP.txt};
            \addplot[blue] table[x=t, y=b] {figures/data/Window_SMAP.txt};
            \addplot[gray, dashed] coordinates {(4,0.10)(4,0.70)};
        \end{axis}
    \end{tikzpicture}
    \caption{\textit{SMAP}} \label{subfig:WindowSMAP}  
    \end{subfigure}
\caption{{Hyperparameter selection---the numbers over the \textit{PR} curve are candidate values for window size $w$.}
} 
\label{fig:window_size}
\end{figure}

We examine the effect of different window sizes. 
%
%
%
%
The results are shown in Figure~\ref{fig:window_size}, where we notice that the window size $w$, using the median strategy is not the best, since other configurations exist with higher PR and ROC values. 
Even so, the median strategy is reliable since the results are balanced among all the evaluated cases. 

\subsubsection{Effect of the number of Basic Models}\label{sssec:accuracy}
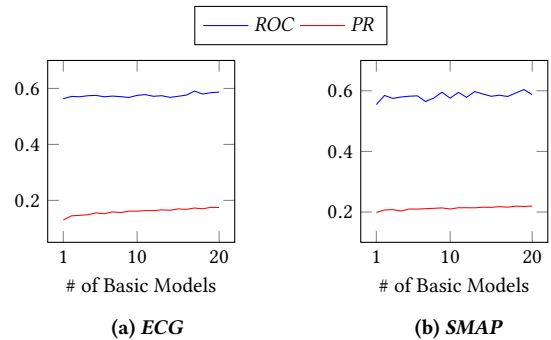
\begin{figure}[ht!] 
\small
\centering 
    \begin{subfigure}[b]{0.48\linewidth}
    \hspace{0.03\linewidth}
        \begin{tikzpicture}
        \begin{axis}[
            width=\linewidth,
            height=0.55*\axisdefaultheight,
            ymax=0.70,
            ymin=0.05,
            xtick={1,10,20},  
            name=bottom axis,
            xlabel=\# of Basic Models,
            legend style={at={(1.3,1.3)},anchor=north,legend columns=-1}]
        ]
            \addplot[blue] table[x=t, y=b] {figures/data/Evolution_ECG_All.txt};
            \addlegendentry{\textit{ROC}}
            \addplot[red] table[x=t, y=a] {figures/data/Evolution_ECG_All.txt};
            \addlegendentry{\textit{PR}}
        \end{axis}

    \end{tikzpicture}
    \caption{\textit{ECG}} \label{subfig:EnsembleECG}  
    \end{subfigure} 
    \begin{subfigure}[b]{0.48\linewidth}
    \hspace{0.03\linewidth}
        \begin{tikzpicture}
        \begin{axis}[
            width=\linewidth,
            height=0.55*\axisdefaultheight,
            ymax=0.70,
            ymin=0.10,
            xtick={1,10,20},  
            name=bottom axis,
            xlabel=\# of Basic Models,
        ]
            \addplot[red]  table[x=t, y=a] {figures/data/Evolution_SMAP_All.txt};
            \addplot[blue] table[x=t, y=b] {figures/data/Evolution_SMAP_All.txt};
        \end{axis}
    \end{tikzpicture}
    \caption{\textit{SMAP}} \label{subfig:EnsembleSMAP}  
    \end{subfigure}
\caption{
Effect of the number of basic models.
} 
\label{fig:metricsEvolution}
\end{figure}

Figure~\ref{fig:metricsEvolution} shows the performance for \textit{PR}, and \textit{ROC} as the number of basic models in the ensemble grows during training. 
%
%
For both datasets, there is an increasing performance as we include more basic models. For \textit{ECG}, the results for \textit{PR} are steadier and clearer as the number of basic models is greater. Then, for \textit{ROC}, the metric shows a positive trend, but it is more unstable, as there are sudden changes between cases. For \textit{SMAP}, the results are similar as the number of basic models is growing, with a clear and stable trend in \textit{PR} with some fluctuations in \textit{ROC}. 
This suggests that more basic models often help improve outlier detection accuracy, while training more basic models also needs more time. 

\subsubsection{Run Time} In Table~\ref{table:executionTime}, we report the training time of \texttt{RAE}, \texttt{CAE}, \texttt{RAE-Ensemble}, and \texttt{CAE-Ensemble} on five datasets, with eight basic models for each case. 
%
%
The ensemble models take longer time than their corresponding basic models, which is expected. 
However, \texttt{CAE-Ensemble} is much faster than \texttt{RAE-Ensemble}. For example, in cases such as \textit{SMD}, the training for \texttt{CAE-Ensemble} finishes in a quarter of the time required for \texttt{RAE-Ensemble}. 
The results suggest that \texttt{CAE-Ensemble} is very efficient in comparison to the recurrent autoencoder. 
Also, the runtime ratio of \texttt{RAE-Ensemble} and \texttt{RAE} is on average 7.82, while the ratio between the \texttt{CAE-Ensemble} and \texttt{CAE} is 5.91, indicating that the parameter transfer in the \texttt{CAE-Ensemble} is effective reducing the training time.
Thus, the convolution autoencoder and the parameter transfer are effective and succeed in reducing the running time in comparison to the recurrent model. 

We report the run-time for the online testing in Table~\ref{table:inferenceTime}.
In a streaming setting, we aim at returning an outlier score whenever we receive a new observation. To do so, we create a window with the observation and its previous $w-1$ observations. Then, we use a \texttt{CAE} or \texttt{CAE-Ensemble} to reconstruct the window, which computes the outlier score of the observation, i.e., the last observation in the window (see Section~\ref{sssec:hyperWindows}). 
Table~\ref{table:inferenceTime} shows that each window can be processed in around fifty microseconds, which is efficient enough to support many online streaming settings. 
That is possible because the heavy training computations have already been done offline. In the online phase, a window (e.g., consisting 16 to 32 observations) just needs to go through the computations of \texttt{CAE} or \texttt{CAE-Ensemble} using already learned parameters, which is very efficient.   
\texttt{CAE-Ensemble} is only slightly slower than \texttt{CAE} as the multiple basic models in \texttt{CAE-Ensemble} can run in parallel. 

\begin{table}[ht!]
\small
    \centering
    \caption{Training Time Comparison (in minutes).}
    \label{table:executionTime}
    \begin{tabular}{ |l|c|c|c|c|c| } 
        \hline
        \textbf{Model} &  \textit{ECG} & \textit{MSL} & \textit{SMAP} & \textit{SMD} & \textit{WADI}\\ 
        \hline
\texttt{RAE} & 7.84 & 16.63 & 32.19 & 246.43 & 72.32 \\
\texttt{RAE-Ensemble} & 59.66 & 129.99 & 254.83 & 1959.13 & 566.89 \\
\hhline{|~|-|-|-|-|-|}
\texttt{Ratio} & 7.60 & 7.82 & 7.92 & 7.95 & 7.84 \\
        \hline
\texttt{CAE} & 4.16 & 7.65 & 20.36 & 74.34 & 22.37 \\
\texttt{CAE-Ensemble} & 24.05 & 45.45 & 122.13 & 452.06 & 129.58 \\
\hhline{|~|-|-|-|-|-|}
\texttt{Ratio} & 5.78 & 5.94 & 6.00 & 6.08 & 5.79 \\
        \hline
    \end{tabular}
\end{table}

\begin{table}[ht!]
\small
    \centering
    \caption{Inference Time per Window (milliseconds).}
    \label{table:inferenceTime}
    \begin{tabular}{ |l|c|c|c|c|c| } 
        \hline
        \textbf{} &  \textit{ECG} & \textit{MSL} & \textit{SMAP} & \textit{SMD} & \textit{WADI}\\ 
        \hline
        \texttt{CAE} & 0.0489 & 0.0517 & 0.0500 & 0.0465 & 0.0546 \\
        \texttt{CAE-Ensemble} & 0.0499 & 0.0520 & 0.0505 & 0.0469 & 0.0549 \\
        \hline
    \end{tabular}
    \vspace{-1em}
\end{table}

\subsubsection{Kernel Size} 
\begin{figure}[ht!] 
\small
\centering 

    \begin{subfigure}[t]{0.48\linewidth}
    \hspace{0.03\linewidth}
        \begin{tikzpicture}
        \begin{axis}[
            width=\linewidth,
            height=0.55*\axisdefaultheight,
            ymax=0.70,
            ymin=0.10,
            xtick={3,5,7,9},
            name=bottom axis,
            xlabel=Kernel size,
            legend style={at={(1.3,1.4)},anchor=north,legend columns=-1}]
        ]
            \addplot[brown] table[x=t, y=pre] {figures/data/Kernel_ECG.txt};
            \addlegendentry{\textit{Precision}}
            \addplot[purple] table[x=t, y=re] {figures/data/Kernel_ECG.txt};
            \addlegendentry{\textit{Recall}}
            \addplot[green] table[x=t, y=f1] {figures/data/Kernel_ECG.txt};
            \addlegendentry{\textit{F1}}
            \addplot[red] table[x=t, y=pr] {figures/data/Kernel_ECG.txt};
            \addlegendentry{\textit{PR}}
            \addplot[blue] table[x=t, y=roc] {figures/data/Kernel_ECG.txt};
            \addlegendentry{\textit{ROC}}
        \end{axis}

    \end{tikzpicture}
    \caption{\textit{ECG}} \label{subfig:kernelECG}  
    \end{subfigure}
    \begin{subfigure}[t]{0.48\linewidth}
    \hspace{0.03\linewidth}
        \begin{tikzpicture}
        \begin{axis}[
            width=\linewidth,
            xtick={3,5,7,9},
            height=0.55*\axisdefaultheight,
            name=bottom axis,
            xlabel=Kernel size,
        ]
            \addplot[brown] table[x=t, y=pre] {figures/data/Kernel_SMAP.txt};
            \addplot[green] table[x=t, y=f1] {figures/data/Kernel_SMAP.txt};
            \addplot[purple] table[x=t, y=re] {figures/data/Kernel_SMAP.txt};
            \addplot[red] table[x=t, y=pr] {figures/data/Kernel_SMAP.txt};
            \addplot[blue] table[x=t, y=roc] {figures/data/Kernel_SMAP.txt};

        \end{axis}
    \end{tikzpicture}
    \caption{\textit{SMAP}} \label{subfig:kernelSMAP}  
    \end{subfigure}

\caption{Effect of kernel size.
}
\label{fig:kernel_size}
\vspace{-1em}
\end{figure}
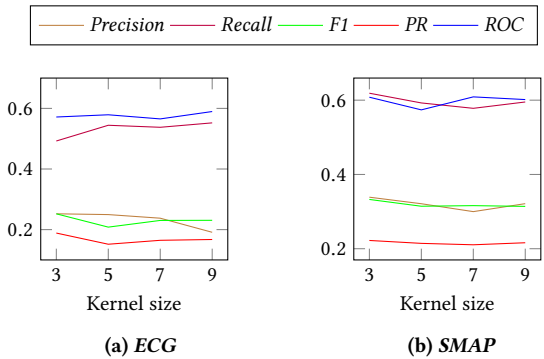

We have conducted an additional experiment by where we vary the kernel size among 3, 5, 7, and 9. The results in Figure~\ref{fig:kernel_size} show that the accuracy is insensitive to the kernel size. 

%
%
%


\section{Related Work} \label{relatedwork}
\textbf{Outlier Detection in Time Series. }
Several traditional methods exist~\cite{Gupta14} that rely mainly on parametric methodologies to detect outliers and further to repair outliers~\cite{ZhangS0Y17}. Thus, they generally achieve good results given proper refinement and tuning according to specific guidelines~\cite{Le20,Zhang17}. 
 %
Some studies
~\cite{TranMS20,YoonLL19} focus on improving the efficiency of distance based outlier detection methods to better support streaming data settings. %
\texttt{CAE-Ensemble} is also able to support streaming settings since the training is performed offline.
Other studies exist that target semi-supervised~\cite{HendrycksMKS19} and supervised settings~\cite{Hundman18}, which require ground-truth labels. 
In contrast, we study unsupervised outlier detection. 
Deep learning based methods are summarized in Table~\ref{table:summary}.


\noindent
\textbf{Sequence-to-sequence (seq2seq) Processing. }
Seq2seq modeling is introduced to model sequences, which is 
intrinsically sequential. 
Recurrent seq2seq architectures 
~\cite{Malhotra16,Hu16,DBLP:conf/icde/Hu0GJX20,tkdesean,MileTS} require expensive 
recursions.
A convolutional architecture 
~\cite{Gehring17} makes it possible to take advantage of parallel execution. 
%
Our framework differs from other seq2seq models by focusing on the unsupervised time series outlier detection problem using the convolution process. To the best of our knowledge, our study is the first attempt to apply seq2seq convolution in unsupervised time series outlier detection. 

\noindent
\textbf{Ensemble Learning. }
Ensembles are used in a broad range of applications, generally covering supervised settings such as classification and regression~\cite{Okun11}. 
Although several ensemble methods exist, such as bagging~\cite{Breiman96b}, boosting~\cite{FreundS97}, and stacking~\cite{Wolpert92}, these methods all depend on labeled data to aggregate basic models. 
As a result, it is not trivial to apply these methods for unsupervised outlier detection. 
Existing unsupervised ensemble studies~\cite{Chen17,Kieu19} form an ensemble by creating a large quantity of basic models with random architectures. However, these methods is sub-optimal because the obtained basic models can be strongly alike thus limiting the diversity.
A recent study~\cite{Zhang20} proposes \texttt{EDGE}, a solution to this problem using a diversity measure among basic models, which inspires our work. However, \texttt{EDGE} targets a supervised setting that differs substantially from our setting of unsupervised outlier detection. In addition, \texttt{EDGE} does not work for sequential data.


\section{Conclusion and future work}
\label{conclusion}
We propose a diversity-driven ensemble model built on convolutional sequence-to-sequence autoencoders for unsupervised outlier detection in time series data, along with an unsupervised hyperparameter selection method. An extensive empirical study offers evidence that the proposal is capable of improving both accuracy and efficiency compared to existing autoencoder based methods. 
%
%

In future work, it is of interest to enable unsupervised time series cleaning by repairing detected outliers. It is also of interest to study more advanced unsupervised hyperparameter selection, e.g., exploring the relationships between the outlier ratio and the diversity metric.


%
\begin{acks}
This work was partially supported by Independent Research Fund
Denmark under agreements 8022-00246B and 8048-00038B, the VILLUM FONDEN under agreement 34328, Huawei Cloud Database Innovation Lab, 
and the Innovation
Fund Denmark centre, DIREC. 
\end{acks}

\clearpage

\onecolumn
\begin{multicols}{2}
\bibliographystyle{unsrt}
\bibliography{References}

\begin{thebibliography}{10}

\bibitem{pvldb22}
David Campos, Tung Kieu, Chenjuan Guo, Feiteng Huang, Kai Zheng, Bin Yang, and
  Christian~S. Jensen.
\newblock Unsupervised time series outlier detection with diversity-driven
  convolutional ensembles.
\newblock {\em Proc. {VLDB} Endow.}, 2022.

\bibitem{DBLP:conf/icde/LiuJYZ18}
Huiping Liu, Cheqing Jin, Bin Yang, and Aoying Zhou.
\newblock Finding top-k optimal sequenced routes.
\newblock In {\em {ICDE}}, pages 569--580, 2018.

\bibitem{Hundman18}
Kyle Hundman, Valentino Constantinou, Christopher Laporte, Ian Colwell, and Tom
  S{\"{o}}derstr{\"{o}}m.
\newblock Detecting spacecraft anomalies using {LSTMs} and nonparametric
  dynamic thresholding.
\newblock In {\em {SIGKDD}}, pages 387--395, 2018.

\bibitem{DBLP:conf/ijcai/YangGHT021}
Sean~Bin Yang, Chenjuan Guo, Jilin Hu, Jian Tang, and Bin Yang.
\newblock Unsupervised path representation learning with curriculum negative
  sampling.
\newblock In {\em {IJCAI}}, pages 3286--3292, 2021.

\bibitem{DBLP:conf/icde/Pedersen0J20}
Simon~Aagaard Pedersen, Bin Yang, and Christian~S. Jensen.
\newblock A hybrid learning approach to stochastic routing.
\newblock In {\em {ICDE}}, pages 1910--1913, 2020.

\bibitem{DBLP:conf/icde/Yang020}
Sean~Bin Yang and Bin Yang.
\newblock Learning to rank paths in spatial networks.
\newblock In {\em {ICDE}}, pages 2006--2009, 2020.

\bibitem{razvanicde2021}
Razvan-Gabriel Cirstea, Tung Kieu, Chenjuan Guo, Bin Yang, and Sinno~Jialin
  Pan.
\newblock Enhancenet: Plugin neural networks for enhancing correlated time
  series forecasting.
\newblock In {\em {ICDE}}, pages 1739--1750, 2021.

\bibitem{DBLP:journals/vldb/HuYGJ18}
Jilin Hu, Bin Yang, Chenjuan Guo, and Christian~S. Jensen.
\newblock Risk-aware path selection with time-varying, uncertain travel costs:
  a time series approach.
\newblock {\em {VLDB} J.}, 27(2):179--200, 2018.

\bibitem{Aggarwal13}
Charu~C. Aggarwal.
\newblock {\em Outlier Analysis}.
\newblock 2013.

\bibitem{DBLP:journals/vldb/PedersenYJ20}
Simon~Aagaard Pedersen, Bin Yang, and Christian~S. Jensen.
\newblock Fast stochastic routing under time-varying uncertainty.
\newblock {\em {VLDB} J.}, 29(4):819--839, 2020.

\bibitem{DBLP:journals/vldb/GuoYHJC20}
Chenjuan Guo, Bin Yang, Jilin Hu, Christian~S. Jensen, and Lu~Chen.
\newblock Context-aware, preference-based vehicle routing.
\newblock {\em {VLDB} J.}, 29(5):1149--1170, 2020.

\bibitem{DBLP:journals/pvldb/PedersenYJ202}
Simon~Aagaard Pedersen, Bin Yang, and Christian~S. Jensen.
\newblock Anytime stochastic routing with hybrid learning.
\newblock {\em Proc. {VLDB} Endow.}, 13(9):1555--1567, 2020.

\bibitem{DBLP:conf/waim/YuanSWYZY10}
Peisen Yuan, Chaofeng Sha, Xiaoling Wang, Bin Yang, Aoying Zhou, and Su~Yang.
\newblock {XML} structural similarity search using mapreduce.
\newblock In {\em {WAIM}}, pages 169--181, 2010.

\bibitem{Gupta14}
Manish Gupta, Jing Gao, Charu~C. Aggarwal, and Jiawei Han.
\newblock Outlier detection for temporal data: {A} survey.
\newblock {\em {IEEE} Trans. Knowl. Data Eng.}, 26(9):2250--2267, 2014.

\bibitem{Hawkins02}
Simon Hawkins, Hongxing He, Graham~J. Williams, and Rohan~A. Baxter.
\newblock Outlier detection using replicator neural networks.
\newblock In {\em {DAWAK}}, pages 170--180, 2002.

\bibitem{DBLP:conf/cikm/Kieu0GJ18}
Tung Kieu, Bin Yang, Chenjuan Guo, and Christian~S. Jensen.
\newblock Distinguishing trajectories from different drivers using incompletely
  labeled trajectories.
\newblock In {\em {CIKM}}, pages 863--872, 2018.

\bibitem{Chen17}
Jinghui Chen, Saket Sathe, Charu~C. Aggarwal, and Deepak~S. Turaga.
\newblock Outlier detection with autoencoder ensembles.
\newblock In {\em {SDM}}, pages 90--98, 2017.

\bibitem{Okun11}
Oleg Okun, Giorgio Valentini, and Matteo R{\'{e}}, editors.
\newblock {\em Ensembles in Machine Learning Applications}, volume 373 of {\em
  Studies in Computational Intelligence}.
\newblock 2011.

\bibitem{Kieu19}
Tung Kieu, Bin Yang, Chenjuan Guo, and Christian~S. Jensen.
\newblock Outlier detection for time series with recurrent autoencoder
  ensembles.
\newblock In {\em {IJCAI}}, pages 2725--2732, 2019.

\bibitem{Malhotra16}
Pankaj Malhotra, Anusha Ramakrishnan, Gaurangi Anand, Lovekesh Vig, Puneet
  Agarwal, and Gautam~M. Shroff.
\newblock {LSTM}-based encoder-decoder for multi-sensor anomaly detection.
\newblock In {\em {ICML} Anomaly Detection Workshop}, page~5, 2016.

\bibitem{Kieu18}
Tung Kieu, Bin Yang, and Christian~S. Jensen.
\newblock Outlier detection for multidimensional time series using deep neural
  networks.
\newblock In {\em {MDM}}, pages 125--134, 2018.

\bibitem{Aggarwal17}
Charu~C. Aggarwal and Saket Sathe.
\newblock {\em Outlier Ensembles - An Introduction}.
\newblock 2017.

\bibitem{LeCun98}
Y.~{Lecun}, L.~{Bottou}, Y.~{Bengio}, and P.~{Haffner}.
\newblock Gradient-based learning applied to document recognition.
\newblock {\em Proceedings of the IEEE}, 86(11):2278--2324, 1998.

\bibitem{Gehring17}
Jonas Gehring, Michael Auli, David Grangier, Denis Yarats, and Yann~N. Dauphin.
\newblock Convolutional sequence to sequence learning.
\newblock In {\em {ICML}}, pages 1243--1252, 2017.

\bibitem{Furlanello18}
Tommaso Furlanello, Zachary~Chase Lipton, Michael Tschannen, Laurent Itti, and
  Anima Anandkumar.
\newblock Born-again neural networks.
\newblock In {\em {ICML}}, pages 1602--1611, 2018.

\bibitem{Sakurada14}
Mayu Sakurada and Takehisa Yairi.
\newblock Anomaly detection using autoencoders with nonlinear dimensionality
  reduction.
\newblock In {\em {MLSDA}}, pages 4--11, 2014.

\bibitem{Hochreiter97}
Sepp Hochreiter and J{\"{u}}rgen Schmidhuber.
\newblock Long short-term memory.
\newblock {\em Neural Comput.}, 9(8):1735--1780, 1997.

\bibitem{Dauphin17}
Yann~N. Dauphin, Angela Fan, Michael Auli, and David Grangier.
\newblock Language modeling with gated convolutional networks.
\newblock In {\em {ICML}}, pages 933--941, 2017.

\bibitem{DBLP:conf/cikm/CirsteaMMG018}
Razvan{-}Gabriel Cirstea, Darius{-}Valer Micu, Gabriel{-}Marcel Muresan,
  Chenjuan Guo, and Bin Yang.
\newblock Correlated time series forecasting using multi-task deep neural
  networks.
\newblock In {\em {CIKM}}, pages 1527--1530, 2018.

\bibitem{Sammut17}
Claude Sammut and Geoffrey~I. Webb, editors.
\newblock {\em Encyclopedia of Machine Learning and Data Mining}.
\newblock 2017.

\bibitem{Hu16}
Renjun Hu, Charu~C. Aggarwal, Shuai Ma, and Jinpeng Huai.
\newblock An embedding approach to anomaly detection.
\newblock In {\em {ICDE}}, pages 385--396, 2016.

\bibitem{Vaswani17}
Ashish Vaswani, Noam Shazeer, Niki Parmar, Jakob Uszkoreit, Llion Jones,
  Aidan~N. Gomez, Lukasz Kaiser, and Illia Polosukhin.
\newblock Attention is all you need.
\newblock In {\em {NIPS}}, pages 5998--6008, 2017.

\bibitem{He16}
Kaiming He, Xiangyu Zhang, Shaoqing Ren, and Jian Sun.
\newblock {Deep Residual Learning for Image Recognition}.
\newblock In {\em {CVPR}}, pages 770--778, 2016.

\bibitem{Luong15}
Thang Luong, Hieu Pham, and Christopher~D. Manning.
\newblock Effective approaches to attention-based neural machine translation.
\newblock In {\em {EMNLP}}, pages 1412--1421, 2015.

\bibitem{Wang10}
Shuo Wang, Huanhuan Chen, and Xin Yao.
\newblock Negative correlation learning for classification ensembles.
\newblock In {\em {IJCNN}}, pages 1--8, 2010.

\bibitem{Huang17}
Gao Huang, Yixuan Li, Geoff Pleiss, Zhuang Liu, John~E. Hopcroft, and Kilian~Q.
  Weinberger.
\newblock Snapshot ensembles: Train 1, get {M} for free.
\newblock In {\em {ICLR}}, page pp. 14, 2017.

\bibitem{Zhang20}
Wentao Zhang, Jiawei Jiang, Yingxia Shao, and Bin Cui.
\newblock Efficient diversity-driven ensemble for deep neural networks.
\newblock In {\em {ICDE}}, pages 73--84, 2020.

\bibitem{Cevikalp20}
Hakan Cevikalp, Burak Benligiray, and Omer~Nezih Gerek.
\newblock {Semi-supervised Robust Deep Neural Networks for Multi-label Image
  Classification}.
\newblock {\em Pattern Recognition}, 100:107164, 2020.

\bibitem{BergstraB12}
James Bergstra and Yoshua Bengio.
\newblock Random search for hyper-parameter optimization.
\newblock {\em J. Mach. Learn. Res.}, 13:281--305, 2012.

\bibitem{Liu08}
Fei~Tony Liu, Kai~Ming Ting, and Zhi{-}Hua Zhou.
\newblock Isolation forest.
\newblock In {\em {ICDM}}, pages 413--422, 2008.

\bibitem{Breunig00}
Markus~M. Breunig, Hans{-}Peter Kriegel, Raymond~T. Ng, and J{\"{o}}rg Sander.
\newblock {LOF:} identifying density-based local outliers.
\newblock In {\em {SIGMOD}}, pages 93--104, 2000.

\bibitem{Scholkopf99}
Bernhard Sch{\"{o}}lkopf, Robert~C. Williamson, Alexander~J. Smola, John
  Shawe{-}Taylor, and John~C. Platt.
\newblock Support vector method for novelty detection.
\newblock In {\em {NIPS}}, pages 582--588, 1999.

\bibitem{ScholkopfSWB00}
Bernhard Sch{\"{o}}lkopf, Alexander~J. Smola, Robert~C. Williamson, and
  Peter~L. Bartlett.
\newblock New support vector algorithms.
\newblock {\em Neural Comput.}, 12(5):1207--1245, 2000.

\bibitem{Zhang19}
Chuxu Zhang, Dongjin Song, Yuncong Chen, Xinyang Feng, Cristian Lumezanu, Wei
  Cheng, Jingchao Ni, Bo~Zong, Haifeng Chen, and Nitesh~V. Chawla.
\newblock A deep neural network for unsupervised anomaly detection and
  diagnosis in multivariate time series data.
\newblock In {\em {AAAI}}, pages 1409--1416, 2019.

\bibitem{Soelch16}
Maximilian Soelch, Justin Bayer, Marvin Ludersdorfer, and Patrick van~der
  Smagt.
\newblock Variational inference for on-line anomaly detection in
  high-dimensional time series.
\newblock {\em CoRR}, abs/1602.07109:4, 2016.

\bibitem{Su19}
Ya~Su, Youjian Zhao, Chenhao Niu, Rong Liu, Wei Sun, and Dan Pei.
\newblock Robust anomaly detection for multivariate time series through
  stochastic recurrent neural network.
\newblock In {\em {SIGKDD}}, pages 2828--2837, 2019.

\bibitem{Xu18}
Haowen Xu, Wenxiao Chen, Nengwen Zhao, Zeyan Li, Jiahao Bu, Zhihan Li, Ying
  Liu, Youjian Zhao, Dan Pei, Yang Feng, Jie Chen, Zhaogang Wang, and Honglin
  Qiao.
\newblock Unsupervised anomaly detection via variational auto-encoder for
  seasonal {KPI}s in web applications.
\newblock In {\em {WWW}}, pages 187--196, 2018.

\bibitem{Kingma15}
Diederik~P. Kingma and Jimmy Ba.
\newblock Adam: a method for stochastic optimization.
\newblock In {\em {ICLR}}, page pp. 15, 2015.

\bibitem{ZhangS0Y17}
Aoqian Zhang, Shaoxu Song, Jianmin Wang, and Philip~S. Yu.
\newblock Time series data cleaning: From anomaly detection to anomaly
  repairing.
\newblock {\em Proc. {VLDB} Endow.}, 10(10):1046--1057, 2017.

\bibitem{Le20}
Kim{-}Hung Le and Paolo Papotti.
\newblock User-driven error detection for time series with events.
\newblock In {\em {ICDE}}, pages 745--757, 2020.

\bibitem{Zhang17}
Xuyun Zhang, Wan{-}Chun Dou, Qiang He, Rui Zhou, Christopher Leckie, Kotagiri
  Ramamohanarao, and Zoran~A. Salcic.
\newblock Lshiforest: {A} generic framework for fast tree isolation based
  ensemble anomaly analysis.
\newblock In {\em {ICDE}}, pages 983--994, 2017.

\bibitem{TranMS20}
Luan Tran, Minyoung Mun, and Cyrus Shahabi.
\newblock Real-time distance-based outlier detection in data streams.
\newblock {\em Proc. {VLDB} Endow.}, 14(2):141--153, 2020.

\bibitem{YoonLL19}
Susik Yoon, Jae{-}Gil Lee, and Byung~Suk Lee.
\newblock {NETS:} extremely fast outlier detection from a data stream via
  set-based processing.
\newblock {\em Proc. {VLDB} Endow.}, 12(11):1303--1315, 2019.

\bibitem{HendrycksMKS19}
Dan Hendrycks, Mantas Mazeika, Saurav Kadavath, and Dawn Song.
\newblock Using self-supervised learning can improve model robustness and
  uncertainty.
\newblock In {\em {NIPS}}, pages 15637--15648, 2019.

\bibitem{DBLP:conf/icde/Hu0GJX20}
Jilin Hu, Bin Yang, Chenjuan Guo, Christian~S. Jensen, and Hui Xiong.
\newblock Stochastic origin-destination matrix forecasting using dual-stage
  graph convolutional, recurrent neural networks.
\newblock In {\em {ICDE}}, pages 1417--1428, 2020.

\bibitem{tkdesean}
Sean~Bin Yang, Chenjuan Guo, and Bin Yang.
\newblock Context-aware path ranking in road networks.
\newblock {\em {IEEE} Trans. Knowl. Data Eng.}, 2020.

\bibitem{MileTS}
Razvan-Gabriel Cirstea, Bin Yang, and Chenjuan Guo.
\newblock Graph attention recurrent neural networks for correlated time series
  forecasting.
\newblock In {\em {MileTS19@KDD}}, 2019.

\bibitem{Breiman96b}
Leo Breiman.
\newblock Bagging predictors.
\newblock {\em Mach. Learn.}, 24(2):123--140, 1996.

\bibitem{FreundS97}
Yoav Freund and Robert~E. Schapire.
\newblock A decision-theoretic generalization of on-line learning and an
  application to boosting.
\newblock {\em J. Comput. Syst. Sci.}, 55(1):119--139, 1997.

\bibitem{Wolpert92}
David~H. Wolpert.
\newblock Stacked generalization.
\newblock {\em Neural Networks}, 5(2):241--259, 1992.

\end{thebibliography}

\end{multicols}
\end{document}